\documentclass[paper=a4, fontsize=13pt]{scrartcl} 
\usepackage{geometry}
\usepackage{tikz}
\usepackage{listings}
\usepackage{algorithmic}
\usepackage{algorithm}
\usepackage{multicol}
\tikzstyle{startstop} = [rectangle, rounded corners, minimum width=3cm, minimum height=1cm,text centered, draw=black, fill=red!30]
\tikzstyle{io} = [trapezium, trapezium left angle=70, trapezium right angle=110, minimum width=3cm, minimum height=1cm, text centered, draw=black, fill=blue!30]
\tikzstyle{process} = [rectangle, minimum width=3cm, minimum height=1cm, text centered, draw=black, fill=orange!30]
\tikzstyle{decision} = [diamond, minimum width=3cm, minimum height=1cm, text centered, draw=black, fill=green!30]
\tikzstyle{arrow} = [thick,->,>=stealth]
\usetikzlibrary{shapes.geometric, arrows}
\usepackage[T1]{fontenc} 
\usepackage{fourier} 
\usepackage[english]{babel} 
\usepackage{amsmath,amsfonts,amsthm} 
\usepackage{lipsum} 

\usepackage{caption}
\usepackage{subcaption}
\usepackage{graphicx}

\usepackage{float}
\usepackage{color}

\usepackage{ulem}

\usepackage{blindtext} 

\usepackage[]{hyperref}  

\usepackage{sectsty} 
\allsectionsfont{\centering \normalfont\scshape} 

\usepackage{fancyhdr} 
\makeatletter
\def\ALG@special@indent{%
    \ifdim\ALG@thistlm=0pt\relax
        \hskip-\leftmargin
    \else
        \hskip\ALG@thistlm
    \fi
}
\newcommand{\Launch}[1]{\item[]\noindent\ALG@special@indent \textbf{Launch:}\ #1}
\newcommand{\EndKernel}{\item[]\noindent\ALG@special@indent \textbf{End Kernel}}
\makeatother
\numberwithin{equation}{section} 
\numberwithin{figure}{section} 
\numberwithin{table}{section} 

\newcommand{\horrule}[1]{\rule{\linewidth}{#1}} 

\title{	
\normalfont \normalsize 
\textsc{Imperial College London, Computing Department} \\ [25pt] 
\horrule{0.5pt} \\[0.4cm] 
\huge Master of Engineering Individual Project Final Report: Safety-Aware Multi-Agent Apprenticeship Learning  \\ 
\horrule{2pt} \\[0.5cm] 
}

\author{ Author : Junchen Zhao  \\
\and
Supervisor : Dr. Francesco Belardinelli}

\date{\normalsize\today} 

\begin{document}

\maketitle 

\newpage
\tableofcontents



\newpage
\section{Introduction}
\textit{People in Supervision of this project: Dr. Belardinelli Francesco, Borja Gonzalez}

\subsection{Project Motivation}
As the rapid development of Artifical Intelligence in the current technology field, Reinforcement Learning  has been proven as a powerful technique that allows autonomous agents to learn optimal behaviors (called policies) in unknown and complex environments through models of rewards and penalization. 

However, in order to make this technique (Reinforcement Learning) work correctly and get the precise reward function, which returns the feedback to the learning agent about when the agent behaves correctly or not, the reward function needs to be thoroughly specified.

As a result, in real-world complex environments, such as autonomous driving, specifying a correct reward function could be one of the hard tasks to tackle for the Reinforcement Learning model designers. To this end, Apprenticeship Learning techniques, in which the agent can infer a reward function from expert behaviors, are of high interest due to the fact that they could result in highly specified reward function efficiently. 

However, for critical tasks such as autonomous driving, we need to critically consider about the safety-related issues, so as to we need to build techniques to automatically check and ensure that the inferred rewards functions and policies resulted from the Reinforcement Learning model fulfill the needed safety requirements of the critical tasks that we have mentioned previously. 

In order to have a well-designed Reinforcement Learning model, which is able to generate the highly-specified reward function satisfying the safety-related considerations, the technique called "Safety-Aware Apprenticeship Learning"  was built in 2018\cite{ZhouLi2018}, which would be introduced in detail in the later sections. 

Although the technique "Safety-Aware Apprenticeship Learning" has been built, it only considers Single-Agent scenario. In the other word, the current "Safety-Aware Apprenticeship Learning" technique can only be applied to one agent running in an isolated environment, a fact which limits the potential implementation of this technique. One of the potential improvements to this technique can be instead of only considering the Single-Agent scenario, we are motivated to consider the Multi-Agent scenario as an extension to this technique. By extending it, the "Safety-Aware Apprenticeship Learning" technique can be applied to multiple agents running in the same environment at the same moment, a fact which increases the utility of this technique. 

The potential implementation of this extended technique in the real world example can be multiple autonomous driving cars running in the same environment meantime with safety-related property checked. 

\subsection{Project Objective}
Our objective of this project is to make the extension based on the technique mentioned in the paper "Safety-Aware Apprenticeship Learning" written by Weichao Zhou and Wenchao Li\cite{ZhouLi2018} to improve the utility and the efficiency of the existing Reinforcement Learning model from Single-Agent Learning framework to Multi-Agent Learning framework.

In the paper "Safety-Aware Apprenticeship Learning" regarding with the Single-Agent scenario, the key latent techniques include (i) the Probabilistic Computational Tree Logic as the way of model's safety-related property checking and (ii) the Inverse Reinforcement Learning:
\begin{enumerate}
\item Probabilistic Computational Tree Logic . as the way of model checking: 
\begin{itemize}
\item According to the paper "Safety-Aware Apprenticeship Learning ", PCTL can be used to verify properties of a stochastic system such as "is the probability that the agent reaches the unsafe area within 10 steps smaller than 5\%". As a result, PCTL allows for probabilistic quantification of properties, a technique which is also called probabilistic Model checking and can be applied to the policy quantification checking process in reinforcement learning. \cite{Han2009}\cite{ZhouLi2018}
\end{itemize}

\item Inverse Reinforcement Learning: 
\begin{itemize}
\item  Essentially the Inverse Reinforcement Learning is a kind of learning from demonstration techniques where the reward function of a Markov Decision Process is unknown to the learning agent. At the same time, the agent has to derive a good policy by observing an expert's demonstrations.\cite{Ng2004}
\end{itemize}
\end{enumerate}

In order to extend the "Safety-Aware Apprenticeship Learning" to the multi-agent scenario, we need to change the key component Markov Decision Process\cite{Bellman1957} used in the Inverse Reinforcement Learning to Markov Game\cite{Hu1999}, which would be discussed in the later sections.

\subsection{My Contribution Conclusion to the Project}
After introducing our project motivation and project objective, at here, I want to conclude my contributions to the project in the following bullet points:
\begin{enumerate}
\item Regarding with the fact that we will add extension to the Inverse Reinforcement Learning model from Single-Agent scenario to a Multi-Agent scenario. My first contribution to this project is considering the case of extracting safe reward functions from expert behaviors in Multi-Agent scenario instead of being from the Single-Agent scenario. 
\item My second contribution is extending the Single-Agent Learning Framework to a Multi-Agent Learning framework and design a novel Learning framework based on the extension in the end.
\item My final contribution to this project is evaluating empirically the performance of my extension to the Single-Agent Inverse Reinforcement Learning framework.


\end{enumerate}


\newpage
\section{Project Prerequisite Knowledge}
\label{chapter2}
\textit{}

In this section, I will give detailed background explanation and introduction about the latent and prerequisite concepts about the Apprenticeship learning for the purpose of understanding our project objectives. I will summarize the bullet points that I will cover in the this section in the following parts:
\begin{enumerate}
\item (\ref{chapter2.1}) Definition of Apprenticeship Learning
\item (\ref{chapter2.2}) Reinforcement Learning Basic(Markov Decision Process)
\item (\ref{chapter2.3}) Single-Agent Reinforcement Learning
\item (\ref{chapter2.4}) General Definition of Markov Game
\item (\ref{chapter2.5}) Inverse Reinforcement Learning Basic
\item (\ref{chapter2.6}) Counterexample Generation and Probabilistic Computational Logic Tree (PCLT) Model Checking in Safety-Aware Apprenticeship Learning
\begin{enumerate}
\item (\ref{chapter2.6.1}) Introduction to the Counterexample Generation and Probabilistic Computational Logic Tree (PCLT) Model Checking
\item (\ref{chapter2.6.2}) Counterexample Generation in DTMC
\item (\ref{chapter2.6.3}) PCTL DTMC Model Checking in Safety-Aware Apprenticeship Learning
\end{enumerate}
\end{enumerate}

\subsection{Definition of Apprenticeship Learning}
\label{chapter2.1}
We consider the formulation of Apprenticeship Learning(AL) by Abbeel and Ng\cite{Abbeel2004}:
\begin{enumerate}
\item The concept of AL is closely related to reinforcement learning (RL) where an agent learns what actions to take in an environment (known as a policy) by maximizing some notion of long-term reward.
\item In AL, however, the agent is not given the reward function, but instead has to first estimate it from a set of expert demonstrations via a technique called inverse reinforcement learning.
\item The formulation assumes that the reward function is expressible as a linear combination of known state features.
\item  An expert demonstrates the task by maximizing this reward function and the agent tries to derive a policy that can match the feature expectations of the expert's demonstrations. Apprenticeship learning can also be viewed as an instance of the class of techniques known as Learning from Demonstration (LfD).
\end{enumerate}
 
As a result, essentially the Apprenticeship Learning is a kind of learning from demonstration techniques where the reward function of a Markov Decision Process is unknown to the learning agent. At the
same time, the agent has to derive a good policy by observing an expert's demonstrations according to the paper "Safety-Aware Apprenticeship Learning" by Weichao Zhou and Wenchao Li.\cite{ZhouLi2018}

\begin{itemize}
\item Unknown Reward Function: We consider the setting where the unknown reward function to the agent in the Markov Decision Process is assumed to be a linear combination of a set of state features.
\end{itemize}

It's possible for someone who can get confused about the definition of the Apprenticeship Learning if he or she lacks of the background in Reinforcement Learning and Inverse Reinforcement Learning. Therefore, I will give detailed explanation about the Reinforcement Learning basic and the Inverse Reinforcement Learning basic in the follow subsections.

\subsection{Reinforcement Basics(Markov Decision Process)}
\label{chapter2.2}

In this section, I will talk about the reinforcement basics and mainly focus on the Markov Decision Process.\newline

The broad definition of Reinforcement Learning can be defined as such below\cite{P2001}:
\begin{itemize}
\item Reinforcement Learning: Reinforcement learning is an area of machine learning concerned with how software agents ought to take actions in an environment in order to maximize some notion of cumulative reward. Reinforcement learning is one of three basic machine learning paradigms, alongside supervised learning and unsupervised learning.
\end{itemize}

Reinforcement Learning was first formally defined in the learning automate model in 1970s.\cite{Narendra1989} In the early time of 1980s, Sutton and Barto developed temporal-difference learning, which is another form of reinforcement learning.\cite{Sutton1998} Further, attention was drawn to reinforcement learning after Watkins and Dayan proposed Q-learning in 1992, which built the connection between reinforcement learning and Markov Decision Process. \cite{Watkins1992}

In reinforcement learning, one of the founding concepts is the Markov Decision Process(MDP). the key of Q-Learning, which is defined in the framework of Markov Decision Process. Markov Decision Process broadly can be defined as a discrete time stochastic control process. It provides a mathematical framework for modeling decision making in situations where outcomes are partly random and partly under the control of a decision maker. \cite{Bellman1957}

However, specifically the MDP can be defined as a finite tuple, which contains five components \({}\)\{$S$, $A$, $P$,  $\gamma$, $s_0$,  $R$\}, composed a process following a set of actions, which are named as \textbf{Policies} $\pi$. I will make the explanation about the meanings of these five terminologies and what the \textbf{Policy} $\pi$ is below: 
\begin{enumerate}
\item $S$ is a finite set of states; 
\item $A$ is a set of actions;
\item $P$ is a transitional probability function describing the probability of transitioning from one state $s$, which belongs to the state set $S$, to another state by taking action $a$, which belongs to the action set $A$;
\item  $R$ is the reward function which maps each state $s$, which belongs to the state set $S$, to a real number indicating the reward of being in state $s$;
\item $s_0$ is the initial state of the MDP which belongs to the state set $S$ as well;
\item $\gamma$ is the discount factor which describes how future rewards attenuate when a sequence of transitions is made;
\item $\pi$ is defined as any mapping from $S$ to $A$.
\end{enumerate}

\subsection{Single-Agent Reinforcement Learning}
\label{chapter2.3}
Now, by having the brief definition of what the Markov Decision Process(MDP) is in the previous section \ref{chapter2.2} Reinforcement Learning Basics, we should have a general idea about what the structure should be. In this section, I will give a more detailed explanation on how the Markov Decision Process is implemented to the Single-Agent Reinforcement Learning. 

 As we have discussed before, a Markov Decision Process (MDP) can be defined as a finite tuple, which contains five components \({}\)\{$S$, $A$, $P$,  $\gamma$, $s_0$,  $R$\}. In a MDP, the objective of the agent is to find a policy $\pi$ so as to maximize the expected sum of discounted rewards. Therefore, the value function $V$ used for finding the policy $\pi$ is shown below\cite{Hu1999}:
 
\begin{equation} \label{eq:reward value single agent}
V(s, \pi) = \sum_{t=0}^{\infty}\gamma* E(r_{t} |\pi, s_0 = s ) 
\end{equation}
where $s_0$ is the initial state, $r_t$ is the reward at time $t$. At the time $t$, the function above can be rewritten as follow:
\begin{equation} \label{eq:reward value single agent rewritten0}
V(s, \pi) = r(s, a_{\pi}) + \gamma*\sum_{s'}p(s'|s, a_{\pi})V(s', \pi)
\end{equation}
where $a_{\pi}$ is the action dictated by policy $\pi$ given initial state $s$. Because it has been proved that there exists an optimal policy $\pi^{*}$ such that for any $s$ $\in$ $S$, the following form of equation would hold:
\begin{equation} \label{eq:reward value single agent rewritten 1*}
V(s, \pi^{*}) = \max_{a}[r(s, a) + \gamma* \sum_{s'}p(s'|s,a)V(s'|\pi^{*}))]
\end{equation}
where $V(s, \pi^{*})$ is called the optimal value for the state s.

If the agent has the direct access to the reward function and state transition function, it can solve for $\pi^{*}$ by iterative search method. However, there would be learning problem exists when the agent doesn't have the access to the reward function or the state transition probabilities. Now the agent has to interact with the environment to find out the its optimal policy. 

The agent can learn the reward function $R$ and the state transition function $P$ and solve for the optimal policy $\pi^{*}$ using the equation \ref{eq:reward value single agent rewritten 1*} above. We are calling this way of finding the optimal policy $\pi^{*}$ as model-based reinforcement learning.

At the same time, the agent can also learn its optimal policy $\pi^{*}$ without having direct access to the reward function $R$ and the state transition probability function $P$. This kind of approach of finding the optimal policy $\pi^{*}$ is called model-free reinforcement learning . One of the model-free reinforcement learning method is Q-Learning.\cite{LittleManl1994}

The basic idea of Q-learning can be defined as the right-hand side of equation \ref{eq:reward value single agent rewritten 1*}:
\begin{equation} \label{eq:Q-learning value single agent rewritten 2*}
Q^{*}(s, a) = r(s, a) + \gamma* \sum_{s'}p(s'|s,a)V(s',\pi^{*}))
\end{equation}

Based on the equation above, $Q^{*}(s,a)$ is the total discounted reward received for single agent by taking action $a$ in state $s$ and then following the optimal policy $\pi^{*}$. The equation always holds for Q-Learning:
\begin{equation} \label{eq Q-learning value single agent rewritten 3*}
V^{*}(s, \pi^{*}) = \max_{a} Q^{*}(s,a)
\end{equation} 
If we know $Q^{*}(s,a)$, then the optimal policy $\pi^{*}$ can be found based on the equation \ref{eq Q-learning value single agent rewritten 3*}, which would always allow the agent to take an action so that the $Q^{*}(s,a)$ can be maximized at any state $s$ for this agent.

In the Q-Learning, the agent starts with arbitrary initial value of $Q^{*}(s,a)$ for all $s$ $\in$ $S$, $a$ $\in$ $A$. At each time $t$, the agent chooses an action and observes its reward, $r_t$. Then, based on the updated reward $r_t$ at each time step $t$, the $Q^{*}(s,a)$ is updated as well by following:

\begin{equation} \label{eq:Q-learning value single agent rewritten 4*}
Q_{t+1}(s, a) = (1-\alpha_{t})*Q_{t}(s,a) + \alpha_{t}*[ \max_{b}Q_{t}(s_{t+1}, b)]
\end{equation}
where $\alpha_{t}$ $\in$ [0,1) is the learning rate, which needs to decay over time in order for learning algorithm to converge. It has been proved that the equation \ref{eq:Q-learning value single agent rewritten 4*} would finally converge to $Q^{*}(s,a)$ under the assumption that all states and actions have been visited infinitely often.\cite{Watkins1992}

\subsection{General Definition of Markov Game}
\label{chapter2.4}
Due to the reason that we are going to improve the current learning model from single agent scenario to multi-agents scenario, Markov Game is needed in this process. In order to understand how it works, I will give general explanation about what Markov Game is in this subsection and in later section \ref{chapter4.1} where I will give more detail about what the Markov Game framework is and how do I implement this into our project.

Basically, Markov Games are the generalization of the Markov Decision Processes(MDPs) to the case of $N$ interacting agents and a Markov Game is defined as $(S, \gamma, A, P, \triangle, r)$ via: \cite{Hu1999}
\begin{enumerate}
\item A set of states $S$, which is the total global joint states of the agents with all state position possibilities
\item $N$ sets of actions $(A_{i})_{i->N}$;
\item The function $P$: $S$ $\times$ $A_{1}$ $\times$ $A_2$ $\times$ $...$ $\times$ $A_{N}$ $\longrightarrow$ $P(S)$ describes the stochastic transition process between states, where $P(S)$ means the set of probability distributions over the set $S$;
\item By giving that we are in state $s^{t}$ at time $t$, and the agent takes actions \{$a_{1}$ ,....., $a_{N}$\}, the state transitions to $s^{t+1}$ with probability $P(s^{t+1} | s^{t}, a_1,.......,a_N)$;
\item By taking the actions, each agent $i$ obtains a bounded reward given by a function $r_i$: $S$ $\times$ $A_{1}$ $\times$ $A_2$ $\times$ $...$ $\times$ $A_{N}$ $\longrightarrow$ $R$;
\item The function $\triangle$ $\in$ $P(S)$ specifies the probability distribution over state space $S$;
\item $\gamma$ $\in$  [0,1) is the discount factor which describes how future rewards attenuate when a sequence of transitions is made;
\end{enumerate}

Now, by giving the basic definition about the Markov Game, then we can use the bold variables without subscription $i$ to denote the concatenation of all variables for all agents: 
\begin{itemize}
\item For example, $a$ denotes actions of all agents and $r$ denotes all rewards in multi-agent setting.
\end{itemize}

Then, we use the subscript $-i$ to denote all agents except for the agent $i$:
\begin{itemize}
\item For example, $(a_{i} , a_{-i})$ represents $(a_1, ...., a_N)$, which is action of all number of $N$ agents.
\end{itemize}

The objective of each agent $i$ in the multi-agent setting is to maximize the its expected return:
\begin{enumerate}
\item The expected return of the agent is defined as: $E_{\pi}$ $[\sum_{t=1}^{N}\gamma^{t}r_{i,t}]$:
\begin{itemize}
\item $r_{i,t}$ is the reward received $t$ steps into the future.
\end{itemize}
\item Each agent in the Markov Game can achieve its own objective by selecting actions through a stochastic policy $\pi_{i}$ : $S$  $\longrightarrow$ $P(A_{i})$. 
\begin{itemize}
\item Then depending on the context, the policies can be Markovian or require additional coordination signals.
\end{itemize}
\item Finally, based on all of the terms that I explained before in this subsection, we can, for each agent $i$, finally further define the expected return for a state-action pair as:
\begin{itemize}
\item $ExpRet_{i}^{\pi_{i}, \pi_{-i}}$ $(s_{t}, a_{t})$ = $E_{s^{t+1:T}, a^{t+1:T}}$ $[\sum_{l \geq t}\gamma^{l-t}r_{i}(s^{l}, a^{l})|s_{t}, a_{t}, \pi]$
\begin{enumerate}
\item $\pi_{i}, \pi_{-i}$: policies of all number of $N$ agents.
\item $T$: total number of steps.
\item $l$: The Future step.
\item $t$: The current step.
\item $s_{t}$: The state at current step $t$.
\item $a_{t}$: The agent action at current step t.
\item $i$: This denotes the current agent.
\end{enumerate}
\end{itemize}
 
\end{enumerate}

\subsection{Inverse Reinforcement Learning Basics}
\label{chapter2.5}

After the introduction about the standard reinforcement learning and the Markov Decision Process, we have generally understood how the reinforcement learning works. Now, I'm going to give introduction and detailed explanation about the Inverse Reinforcement Learning(IRL). 

According to Andrew Ng\cite{Ng2000}, the IRL problem is to find a reward function that can explain observed behavior. By applying the IRL technique, we aim at recovering the reward function \textbf{$R$} in the MDP tuple which we mentioned in the previous subsection MDP \{$S$, $A$, $P$,  $\gamma$, $s_0$,  $R$\} from a set of $m$ trajectories demonstrated by an expert. 

Based on the setting mentioned above in which we'll recover the reward function $R$ in the MDP tuple from a set of m trajectories demonstrated by an expert, we have the IRL from sampled $m$ Monte Carlo trajectories.

We assume that we have the ability to simulate $m$  trajectories ($m_0$ , $m_1$ , $m_2$ , ....) in the Markov Decision Process from the initial state $s_0$ under the optimal policy $\pi^{*}$ or any policy of our choice. For each policy $\pi$ that we will consider, including the optimal policy $\pi^{*}$, we will need a way of estimating the $V^{\pi}$($s_0$) for any setting of the $\alpha$ $s$, where $\alpha$ $s$  is the unknown parameter that we want to "fit" in the linear function approximation. 
\begin{itemize}
\item In order to achieve this goal of estimating the $V^{\pi}$($s_0$), we first execute the $m$ sampled Monte Carlo trajectories under $\pi$.
\item Then, for each $i$ = 1, ... , $d$, if R = $r_{i}$, define $V_{i}^{\pi}$($s_0$) to be the average empirical return that would have been on these $m$ Monte Carlo trajectories. 
\item For example, if we only take $m$ = 1 trajectories, and the trajectory visited the sequence of states ($s_0$, $s1$, ...), then we have the formula below:
\end{itemize}
 \begin{equation} \label{eq:1}
\hat{V_{i}}^{\pi}(s_0) = r_i(s_0) + \gamma \cdot r_i(s_1) + \gamma^{2} \cdot r_i(s_2) + ...
\end{equation} 
As what we have seen above, if we take $m$ number of sampled trajectories, then the $V^{\pi}$($s_0$) will be the average over the empirical returns of $m$ such trajectories. Then for any setting of the $\alpha_i$ $s$, a natural estimate of $V_{i}^{\pi}$($s_0$) is:

\begin{equation} \label{eq:2}
\hat{V_{i}}^{\pi}(s_0) = \alpha_1 \cdot \hat{V_{i}}^{\pi}(s_0) + \alpha_2 \cdot \hat{V_{i}}^{\pi}(s_0) + ........ + \alpha_d \cdot \hat{V_{i}}^{\pi}(s_0)
\end{equation}
By describing the detail about how to recover the reward function $R$ in the MDP tuple from a set of $m$ trajectories by using IRL. We can finally explain the corresponding algorithm in detail. 
\begin{itemize}
\item First, we find the value estimates as described above for the assumed optimal policy $\pi$* that we are given and the random policy that we randomly choose $\pi_1$.
\item The inductive step is as follow:
\begin{enumerate}
\item We have a set of policies \{$\pi_1$, ...., $\pi_k$ \}
\item We want to find a setting of the $\alpha_i$ $s$ so that the resulting reward function can satisfy as follow:
\end{enumerate}
\end{itemize}
\begin{equation} \label{eq:3}
\hat{V_{i}}^{\pi^{*}}(s_0) \geq V^{\pi_i}(s_0) , i , ...., k
\end{equation}
So, until this point, we should have a clear structure about what the inverse reinforcement learning is. I will discuss more details about how Inverse Reinforcement learning is implemented in the Safety-Aware Apprenticeship Learning in the framework of Markov Decision Process and in the framework of Markov Game in the section \ref{chapter3} and section \ref{chapter4}.

\subsection{Counterexample Generation and Probabilistic Computational Logic Tree (PCLT) Model Checking in Safety-Aware Apprenticeship Learning}
\label{chapter2.6}

\subsubsection{ Introduction to the Counterexample Generation and Probabilistic Computational Logic Tree (PCLT) Model Checking}
\label{chapter2.6.1}
In the Safety-Aware Apprenticeship Learning project, one of the key concepts is generating the counterexamples in the Probabilistic Model Checking Process. This concepts would also be mentioned for multiple times in the following sections in order to demonstrate the process of extending the Single-Agent Apprenticeship Learning to Multi-Agent Apprenticeship Learning.

A main strength of model checking is the possibility of generating the counterexamples in case a property is violated, and it's the most important part in model checking. First of all, the counterexamples provide diagnostic feedback about the model even when there is only a small fragment of the model can be searched. Second, the counterexamples are at the core of obtaining the feasible schedules in timed model checking. The shape of a counterexample depends on the checked formula and the temporal logic. In our project, we are mainly focusing on how the counterexamples are generated in the probabilistic model checking. However, in order to understand how the counterexamples are generated in the probabilistic model checking, we need to first understand what the probabilistic model checking is.

Probabilistic model checking is a technique to verify the system models where transitions in the models equip the random information. The popular system models that probabilistic model checking can be used are the Discrete and Continuous-time Markov Chain (DTMC and CTMC).   Efficient model-checking algorithms for these models have been developed, implemented in a variety of software tools, and applied to case studies from various application areas ranging from randomized distributed algorithms, computer systems, and security protocols to biological systems and quantum computing. 

The key of probabilistic model checking is to appropriately combine techniques from numerical mathematics and operations research with standard-reachability analysis. In this way, properties such as "the maximal probability to reach a set of goal states by avoiding certain states is at most 0.6" can be automatically checked up to a user-defined precision. 

Markov models comprising millions of states can be checked rather fast by dedicated tools such as PRISM, which we are going to use in our project.However, the counterexamples generation techniques in probabilistic model checking have not been fully developed. 

So, in the paper "Counterexample Generation in Probabilistic Model Checking"\cite{Han2009} came up a setting for generating the counterexamples in probabilistic model checking. The setting is considered in which it has already been established that a certain state refutes a given property and it's considered as probabilistic CTL for DTMC models, because all transitions in the DTMC models have their own transition probabilities. 

In this setting, there is a set of paths that instead of a single path in the DTMC models with probabilistic CTL indicating why a given property is refuted. In order to illustrate it, we first consider the property of the DTMC models in the form that $P_{\leq p}$ ($\Phi$ $\cup^{\leq h}$ $\Psi$), where $\Phi$ and $\Psi$ characterize the set of states, $p$ is the probability lower bound, and $h$ is the bound on the maximal allowed number of steps before reaching the goal. state, such as $\Phi$.

If there is a state $s$ refutes the the property formula in the DTMC models, then the probability of all paths in $s$ satisfy the $\Phi$ $\cup^{\leq h}$ $\Psi$ would be greater than the bound probability $p$. We consider two problems that are aimed to provide useful model diagnostic feedback for this property violation: (i) generating strongest evidences and (ii) smallest counterexamples.

(i) Generating strongest evidences: Strongest evidences are the most probable paths that satisfy the property: $\Phi$ $\cup^{\leq h}$ $\Psi$. The strongest evidences contribute mostly to the property refutation, and as a result, are the most informative. For the bound $h$, if we assume it's going infinite, it's shown that the generating the strongest evidences are equivalent to a standard single-source shortest path problem. If we assume the bound $h$ is going finite, it's shown that the generating the strongest evidences is equivalent to the case of  the constrained shortest path problem, which can be solved in the complexity of $O(hm)$, where $m$ is the number of transitions in the DTMC model. Alternatively, the Viterbi algorithm can be used to generate the strongest evidences with the same complexity $O(hm)$\cite{Han2009}. Because of the property of the strongest evidences that are the most probable paths that satisfy the property of the DTMC model, it's evident that the strongest evidences cannot suffice as true counterexamples as the probability of the strongest evidences lies far below probability bound $p$. 

(ii)Generating the smallest counterexamples: Therefore, in order to generate the true counterexamples in the DTMC models with probabilistic CTL, we consider the way of determining the most probable sub-tree rooted at state $s$. At here, we want to determine the smallest counterexamples, so as to we consider the trees of the smallest counterexamples that exceeds the probability bound $p$. At the same time, if we assume such trees with size $k$ are required to maximally exceed the lower probability bound, no sub-trees should exist of size at most $k$ that exceed p. The problem of generating such smallest counterexamples can be cast as a $k$ shortest paths problem ($k$-SP). The time complexity of computing this tree to generate the smallest counterexamples would be $O(hm + hk log(m/n))$, if we assume there $n$ number of states and $m$ numbers of transitions in the DTMC model. This approach is applicable to probability thresholds with lower bounds in the form of $P_{\geq p}$ ($\Phi$ $\cup^{\leq h}$ $\Psi$), as well as to the logic LTL (Linear Temporal Logic). It is applicable to various other models such as Markov reward models and Markov decision processes (MDPs) once a scheduler for an MDP violating an until-formula is obtained. 

\subsubsection{Counterexample Generation in Discrete-Time Markov Chain (DTMC)}
\label{chapter2.6.2}
After giving the introduction to the counterexamples generation and the DTMC, we are going to dive into the detail of the counterexamples generation in DTMC, which mainly serves for the Single-Agent Apprenticeship Learning as a policy model checker. 

First, let $AP$ denote a fixed, finite set of atomic propositions ranged over by a, b, c, . . . . Then a labelled discrete-time Markov chain (DTMC) $D$ is a triple $(S, P, L)$ with $S$ finite set of states, $P$ : $S$ X$S$ $\rightarrow$ $[0,1]$ is a stochastic matrix, and $L$ : $S$ $\rightarrow$ $2^{AP}$ is a labelling function.\cite{Han2009}

For a DTMC, if $\sum_{s'\in S}$ $P(s,s')$ =1, then we say it's a stochastic. If the $\sum_{s'\in S}$ $P(s,s') \in [0,1)$, then we call the model a fully probabilistic system ($FPS$) and it is sub-stochastic. A state $s$ is absorbing if $P(s,s) = 1$ if $s$ only has a self-loop. A path $\sigma$ in $D$ is a state sequence $s_0$, $s_1$, $s_2$ ... such that for all $i$, $P(s_i, s_{i+1})>0$ where $s_i \in S$. The probability $Pr(\sigma)$ for finite path $\sigma = s_0, s_1, ...., s_n$ is defined as $P(s_0, s_1)\cdot P(s_1,s_2) \cdot \cdot \cdot P(s_{n-1, s_n})$ . For the finite set of the paths $C$, $Pr(C) = \sum_{\sigma \in C}Pr(\sigma)$. And we denote $\sigma[i]$ as the $(i+1)$-st state in $\sigma$.

At here we define two terms' syntax in the PCTL for the illustration purpose: (i) $\Phi$: The state formula. (ii) $\Psi$: The path formula. So, if we have the PCTL formula $P_{\leq p}(\Psi)$ , we have:
\begin{equation} \label{eq:counterexample generation DTMC}
s \nvDash P_{\leq p}(\Psi) , iff:  Pr(\sigma | \sigma[0] = s, \sigma \models \Phi ) > p
\end{equation}

So, $P_{\leq p}(\Psi)$ is refuted by state $s$ whenever the total probability mass of all $\Psi$-paths that start in $s$ exceeds the lower probability bound $p$. This indicates that a counterexample for $P_{\leq p}(\Psi)$ is a set of paths starting in state $s$ and satisfying the path formula $\Psi$. As long as $\Psi$ is a path formula whose validity can be witnessed by finite state sequences, finite paths suffice. \cite{Han2009}

Previously, we have defined two problems that are aimed to provide useful DTMC model diagnostic feedback for property violation: (i) generating strongest evidences and (ii) smallest counterexamples.

As a result, we are going to define the strongest evidence. First, we define a finite path $\sigma$ minimally satisfies the path formula $\Psi$ if it satisfies $\Psi$, but no proper prefix of $\sigma$ does so. Then we have: 
\begin{enumerate}
\item (Definition of Strongest Evidence In DTMC) An evidence for $P_{\leq p}(\Psi)$ in state $s$ is a finite path $\sigma$ that starts in $s$ and minimally satisfies $\Psi$. A strongest evidence is an evidence $\sigma^*$ such that $Pr(\sigma^*) \geq Pr(\sigma)$ for any evidence $\sigma$.\cite{ZhouLi2018}\cite{Han2009}
\end{enumerate}

Then we are going to define the smallest counterexample. The intuition of having the smallest counterexample is the smallest counterexample is mostly exceeding the required probability bound given that it has the smallest number of paths. To compute the strongest evidence and smallest counterexample, the DTMC $D$ is transformed to a weighted digraph $G_D = (V, E, w)$, where  $V$ and $E$ are finite sets of vertices and edges, respectively. $V = S$ and $(v, v') \in E$  iff $P(v,v') > 0$, and $w(v,v') = log(P(v,v')^{-1})$. Multiplication of transition probabilities is thus turned into the addition of edges weight along paths. So we have:
\begin{enumerate}
\item (Definition of smallest counterexample) A counterexample for $P_{\leq p}(\Psi)$ in state $s$ is set $C$ of evidences such that $Pr(C) \geq p$. $C^*$ is the smallest counterexample if $|C*| \leq |C|$ for all counterexamples $C$ and $Pr(C^*) \geq Pr(C')$ for any counterexample $C'$ with $|C'|= |C^*|$ . \cite{Han2009}
\item (Lemma 1) For any path $\sigma$ from $s$ to $t$ in DTMC $D$, $k$ $\in$  $N_{>0}$ , and $h$  $\in$ $N$ $\cup$ $\{\infty\}: \sigma$ is a $k$-th most probable path of at most $h$ hops in $D$ iff $\sigma$ is the $k$-th shortest path of at most $h$ hops in $G_D$. \cite{Han2009}
\end{enumerate}

Consider the property formula $\Phi$ $\cup^{\leq h}$ $\Psi$. If state $s$ $\nvDash$ $P_{\leq p}$ $\Phi$ $\cup^{\leq h}$ $\Psi$, then a strongest evidence can be found by a shortest path algorithm (SP algorithm) once all $\Phi$-states and all ($\neg \Psi$ and $\neg \Phi$ )-states in DTMC $D$ are made absorbing. 

Also, the smallest counterexample can be found by applying the $k$-SP algorithms that allow $k$ to be determined, where $k$ is the most probable path of at most $h$ hops in $D$. If $h \neq \infty$, hop-constrained SP with time-complexity $O(hm)$ and $k$-SP algorithms with time complexity $O(hm + hk log(m/n))$ need to be employed, where $n=|S|$ and $m$ is the number of non-zero entries in $P$. 

\subsubsection{PCTL DTMC Model Checking in Safety-Aware Apprenticeship Learning}
\label{chapter2.6.3}
According to the paper "Safety-Aware Apprenticeship Learning " \cite{ZhouLi2018}, PCTL can be used to verify properties of a stochastic system such as "is the probability that the agent reaches the unsafe area within 10 steps smaller than 5\%" in the DTMC models. As a result, PCTL allows for probabilistic quantification of properties, a technique which is also called probabilistic Model checking and can be applied to the policy quantification checking process in reinforcement learning. 

In PCTL for DTMC models, there are two main syntax, including the (i) State Formulas and the (ii) Path Formulas. 

First, let's understand what the State Formulas syntax is.
\begin{enumerate}
\item Generally, we use symbol $\Phi$ to represent the State Formulas.
\item State Formulas asserts the property of a single state $s$ $\in$ $S$  in the MDP.
\item $\Phi$ ::= true |$l_i$|$\neg$ $\Phi_i$ |$\Phi_i$ $\land$ $\Phi_j$|$P_{\triangleright \triangleleft p*}$($\Psi$).
\item $\triangleright$ $\triangleleft$ $\in$ \{$\leq$, $\geq$, $>$, $<$ \}.
\item $P_{\triangleright \triangleleft p*}$($\Psi$) means that the probability of generating a trajectory that satisfies the formulas $\Psi$, which is the Path formulas and we will talk about it in below, is $\triangleright$ $\triangleleft$ $p$*.
\end{enumerate}

Second, let's understand what the Path formulas syntax is.
\begin{enumerate}
\item Generally, we use symbol $\Psi$ to represent the Path formulas.
\item Path Formulas asserts the property of a trajectory.
\item $\Psi$  ::= $X$ $\Phi$ |$\Phi_1$ $\cup^{\leq k} $ $\Phi_2$ |$\Phi_1$ $\cup$ $\Phi_2$ .
\item  $X$ $\Phi$ asserts that the next state after initial state in the trajectory satisfies $\Phi$.
\item $\Phi_1$ $\cup^{\leq k} $ $\Phi_2$ asserts that $\Phi_2$ is satisfied in at most $k$ transitions and all preceding states satisfy $\Phi_1$.
\item $\Phi_1$ $\cup$ $\Phi_2$ asserts that  $\Phi_2$ will be eventually satisfied and all preceding states satisfy $\Phi_1$.
\item The semantics of PCTL is defined by a satisfaction relation $\models$ as follows.
\begin{itemize}
\item s $\models$ true $iff$  state $s$ $\in$ $S$ .
\item s $\models$ $\Phi$  $iff$ state s satisfies the State formula $\Phi$.
\item $\tau$ $\models$ $\Psi$ $iff$ trajectory $\tau$ satisfies the Path formula $\Psi$ .
\end{itemize}
\end{enumerate}

After briefly talking about what the syntax of PCTL is, we can make more detailed explanation about how the PCTL DTMC Model Checking is implemented to the Apprenticeship Learning process as a model checker on the policy in section \ref{chapter3} and section \ref{chapter4}.


\newpage
\section{Single-Agent Safety-Aware Apprenticeship Learning Explanation}
\label{chapter3}


By giving the detailed introduction about prerequisite knowledge for understanding our project in the previous section. In this chapter, I will put the main focus on how the \textbf{Single-agent Safety-Aware Apprenticeship Learning} works due to the reason that our objective main objective is extending the single agent learning system framework to multi-agent learning system framework. I will summarize the bullet points that I will cover in the this section in the following parts:
\begin{enumerate}
\item (\ref{chapter3.1}) Single-Agent Apprenticeship Learning via Inverse Reinforcement Learning.
\begin{enumerate}
\item (\ref{chapter3.1.1}1) Single-Agent Apprenticeship Learning via Inverse Reinforcement Learning Assumption.
\item (\ref{chapter3.1.2}) Optimal Policy Generation Algorithm Notation Demonstration by following Single-Agent Apprenticeship Learning Assumption.
\item (\ref{chapter3.1.3}.3) Optimal Policy Generation Algorithm Demonstration by following Single-Agent Apprenticeship Learning Assumption.
\end{enumerate}
\item (\ref{chapter3.2}) PCTL Model Checking in Single-Agent Apprenticeship Learning.
\item (\ref{chapter3.3}) The Framework for Single-Agent Safety-aware Apprenticeship Learning.
\item (\ref{chapter3.4}) Problem Solved by Single-Agent Safety-Aware Apprenticeship Learning
\end{enumerate}

\subsection{Single-Agent Apprenticeship Learning via Inverse Reinforcement Learning}
\label{chapter3.1}
Because we have introduced in section two about the Inverse Reinforcement Learning IRL, which is finding a reward function that can explain observed behavior and essentially recovering the reward function $R$ which corresponds with the optimal policy $\pi^{*}$ in the $MDP$, we can explain how the $IRL$ is applied to the Apprenticeship learning in this subsection. \cite{Ng2004}

\subsubsection{Single-Agent Apprenticeship Learning via Inverse Reinforcement Learning Assumption.}
\label{chapter3.1.1}
Inverse Reinforcement Learning(IRL) aims at recovering the reward function $R$ of MDP \({}\)\{$S$, $A$, $P$,  $\gamma$, $s_0$,  $R$\} from a set of $m$ trajectories  $\tau$ = \{$\tau_1$ , $\tau_2$ , $\tau_3$, ...\} demonstrated by the experts, where each trajectory $\tau$ is defined as $\tau$ = $\{(s^t, a^t)\}_{t=1}^{T}$, where $t$ represents the iteration time that $t$ $\in$ $T$.

In order to achieve $AL$ via $IRL$, $AL$ assumes that the reward function $R$ of MDP is linear combination of state features. such as $R(s)$ = $w^T$ $f(s)$.\cite{ZhouLi2018}
\begin{enumerate}
\item State features $f(s)$ $\implies$ $[0, 1]$ is a vector of known features over states $S$.
\item $w$ $\in$ $R$ is an unknown weight vector that satisfies $||w_2||$ $\leq$ 1.
\end{enumerate}

\subsubsection{Optimal Policy Generation Algorithm Notation Demonstration by following Single-Agent Apprenticeship Learning Assumption.}
\label{chapter3.1.2}
By knowing the $AL$ assumptions and following the assumption showing above, we can estimate the expected features of a policy $\pi$, the expected features which are expected values of the cumulative discounted state features  $f(s)$ by following $\pi$ on $M$, such that $\mu_E$ $=$ $E$ $[\sum_{t=0}^{\infty} $ $\gamma^t$ $f(s_t)$ $|$ $\pi$ ].
\begin{enumerate}
\item $\mu_E$ denotes the expected features of the unknown expert's policy $\pi_E$.
\item $\gamma^{t}$ denotes the unknown weight vector satisfies  $||\gamma^{t}_2||$ $\leq$ 1 at time $t$.
\item $f$ : state feature $s_t$ $\implies$ $[0, 1]$ is a vector of known features over State $S$.
\item $\mu_E$ can be approximated by the expected features of expert's $m$ demonstrated trajectories: $\mu_E$ $=$ $1/m$ $\sum_{\tau \in \tau_E}$ $\sum_{t=0}^{\infty} $ $\gamma^t$ $f(s_t)$, if the set of demonstrated trajectories by the expert in the size of $m$ is big enough.
\end{enumerate}

As a result, given a error bound $\epsilon$, a policy $\pi^*$ is defined to be $\epsilon$-close to the unknown expert's policy $\pi_E$:
\begin{itemize}
\item If the \textbf{expected feature} $\mu_{\pi^*}$ satisfies the relation that: $||\mu_E - \mu_{\pi^*}||_2$ $\leq$ $\epsilon$.
\item The expected features of the policy $\mu_{\pi^*}$ can be calculated by the Monte Carlo Method, value iteration or Linear Programming.
\end{itemize}

\subsubsection{Optimal Policy Generation Algorithm Demonstration by following Single-Agent Apprenticeship Learning Assumption.}
\label{chapter3.1.3}
In order to calculate the expected features of a policy $\pi^*$and find the optimal policy $\pi^{*}$, we are going to use the algorithm proposed by Abbeel and Ng starts with a random policy $\pi_0$ and its expected policy $\mu_{\pi_0}$.
\begin{enumerate}
\item Assuming in iteration i, we have found a set $i$ candidate policies $\Pi$ = \{${\pi_0, \pi_1, \pi_2, ....}$\} and the corresponding expected features \{$\mu_{\pi}$ $|$ $\pi$ $\in$ $\Pi$\}, then by applying the mini-max algorithm, we have:
\end{enumerate}
\begin{equation} \label{maxmin}
\delta = \max_{w} \min_{\pi \in \Pi} w^T(\hat{\mu_E} - \mu_{\pi})       
s.t. ||w||_{2} \leq 1
\end{equation}
\begin{itemize}
\item The optimal $w$, the unknown weight, is used to find the corresponding optimal policy $\pi_{i}$ and the expected features $\mu_{\pi_{i}}$.
\item If $\delta$ $<$ $\epsilon$, which is the error bound between the expected feature from the unknown expert's policy $\pi_E$ and the current policy $\pi_i$:
\begin{enumerate}
\item The algorithm terminates, and the policy $\pi_i$ is produced as the optimal policy.
\item Otherwise, the expected feature from the current expert's policy, which is $\mu_{\pi_{i}}$is added to the set of features for the policy set $\Pi$ and the algorithm continues the iteration until the optimal policy is found.
\end{enumerate}
\end{itemize}

\subsection{PCTL Model Checking in Single-Agent Apprenticeship Learning}
\label{chapter3.2}
In the previous section \ref{chapter2.6}, I have introduced how the PCTL works as a way of model checking, so I will assume you have familiarized with the terminologies and the concepts for understanding the following contents. Now, I'm going to explain how the PCTL Model Checking works in Apprenticeship Learning.

Based on the Zhou and Li's algorithm for model checking\cite{ZhouLi2018}, they define the $pref(\tau)$ as the set of all $prefixes$  of trajectory $\tau$ including $\tau$ itself, then $\tau$ $\models_{min}$ $\Psi$ ( $\models_{min}$ means there is minimal satisfaction relationship exists) \textbf{\textit{iff}} 

\begin{equation} \label{minimal satisfaction}
(\tau \models \Psi) \land (\forall \tau^{'} \in  pref(\tau) \setminus \tau , \tau^{'} \nvDash \Psi)
\end{equation}
\begin{itemize}
\item In a easier way to explain this satisfaction relationship, we can utilize an example as such:
\begin{enumerate}
\item if $\Psi$ $=$ $\Phi_1$ $\cup^{\leq k}$ $\Phi_2$, then for any finite trajectory, we have the minimal satisfaction relationship exists that:
\begin{itemize}
\item $\tau$ $\models_{min}$  $\Phi_1$ $\cup^{\leq k}$ $\Phi_2$, and only the final state in $\tau$ satisfies $\Phi_2$.
\end{itemize}
\end{enumerate}
\end{itemize}

Therefore, let $P(\tau)$ be the Probability of transitioning along the trajectory $\tau$ and let $\tau_{\psi}$ be the set of all finite trajectories that satisfies $\tau$ $\models_{min}$ $\Psi$ (This relationship is explained above), then the value of PCTL property $\Psi$ is defined as $\sum_{\tau \in \tau_{\Psi}}$ $P(\tau)$.\cite{Hansson1994}
\begin{itemize}
\item So, for a Discrete-time-Markov-Chain (DTMC) $M_{\pi}$ and a state formula, we have:
\begin{enumerate}
\item A \textit{counterexample} of $\Phi$ is a set $cex$ $\subseteq$ $\tau_E$ that satisfies $\sum_{\tau \in cex}$ $P(\tau)$ $>$ $p^{*}$.
\item $P(\tau)$ = $\sum_{\tau^{*} \in \tau}$ $P(\tau^{*})$ is the sum of probability of all trajectories in trajectory set $\tau$.
\item $CEX_{\Phi}$ $\subseteq$ $2^{\tau_{\Psi}}$ is the set of all \textit{counterexamples} for a formula $\Phi$.
\end{enumerate}
\end{itemize}

\subsection{The Framework for Single-Agent Safety-aware Apprenticeship Learning}
\label{chapter3.3}
Based on Zhou and Li\cite{ZhouLi2018}, the framework for safety-aware Apprenticeship Learning can be concluded to the process shown in the figure \textbf{(3.1)}.
Based on this figure, we can generalize the framework as such simplified version in text below:
\begin{enumerate}
\item We utilize Information from both of $Verifier$ and $Expert Demonstration$.
\begin{enumerate}
\item Performing the Model Checking: $Verifier$ check the candidate policy $\pi^{*}$ satisfies the State Formula $\Phi$ or not.
\item If candidate policy $\pi^{*}$ satisfies the State Formula $\Phi$, then:
\begin{enumerate}
\item Check whether our learning objective is met. Our learning objective is to check whether the $\delta$ < $\epsilon$, where $\delta$ is the optimal difference between the expected feature from expert demonstration policy and the expected feature from the current policy, and $\epsilon$ is the error bound that we defined previously.
\begin{enumerate}
\item If we meet our learning objective, then the optimal policy is generated.
\item Otherwise, we add the current candidate policy to the policy set $\Pi$.
\end{enumerate}
\end{enumerate}
\item If candidate policy $\pi^{*}$ doesn't satisfy the State Formula $\Phi$, then:
\begin{enumerate}
\item We generate the counterexample $cex$.
\item Continuing the iteration.
\end{enumerate}
\item The iteration will continue unless the optimal policy is found.
\end{enumerate}
\end{enumerate}

\begin{figure}[h!]
	\centering
	\includegraphics[width=0.6\linewidth]{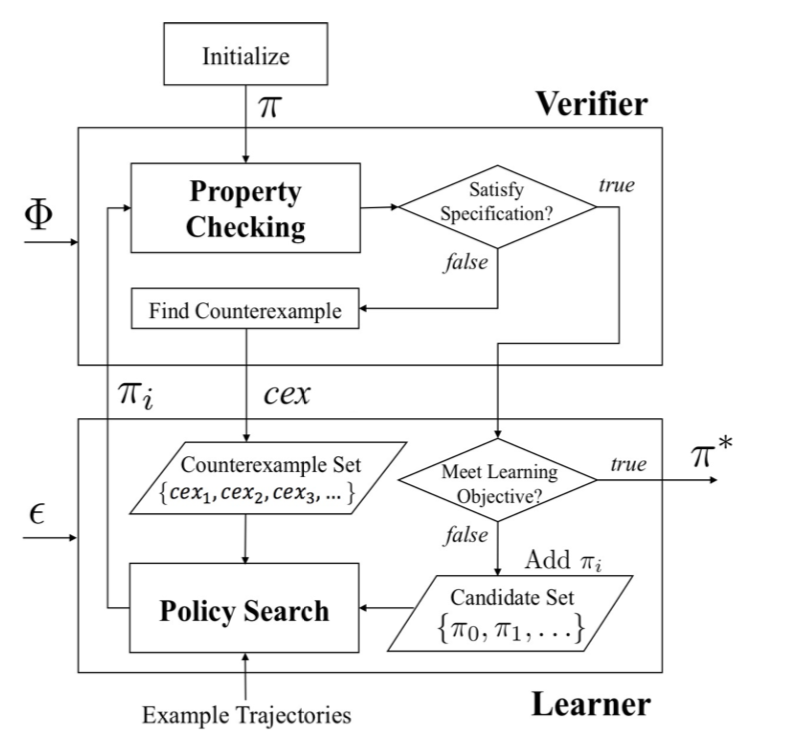}
	\caption{Single-Agent Apprenticeship Learning Framework}
	\label{fig:Framework for Safety-aware Apprenticeship Learning}
\end{figure}

\subsection{Problem Solved by Single-Agent Safety-Aware Apprenticeship Learning}
\label{chapter3.4}
\begin{enumerate}
\item Definition of safety issue in apprenticeship learning: An agent following the learnt policy would have a higher probability of entering those unsafe states than it should.
\item Reasons of having the safety issue in apprenticeship learning:
\begin{enumerate}
\item Expert policy $\pi_{E}$ itself has a high probability of reaching the unsafe states.
\item Human expert often tend to perform only successful demonstrations that do not highlight the unwanted situations. (Lack of negative examples)
\end{enumerate}
\item The safety-aware apprenticeship learning problem :
\begin{enumerate}
\item Given an $MDP$, a set of $m$ trajectories demonstrated by an expert, and a \textbf{specification ($\Phi$)}, to learning a policy that satisfies the state formula and is $error-closed$ to the expert policy $\pi_{E}$.
\end{enumerate}
\end{enumerate}

\newpage
\section{Multi-Agent Safety-Aware Apprenticeship Learning Explanation}
\label{chapter4}
\textit{}


In this section, I will give detailed explanation about Multi-Agent Safety-Aware Apprenticeship Learning, which is mainly focused on the extension from Single-agent environment (MDP) to Multi-agent environment (Markov Game). The procedure of the Multi-Agent Safety-Aware Apprenticeship Learning can be briefly generalized as (i) Extending the game environment from MDP to Markov Game, (ii) Learning the decision rule which contains the joint polices of multiple agents from Multi-Agent Apprenticeship Learning Algorithm, and (iii) Applying safety specification property checker to avoid the learnt decision rule from entering unsafe area. 

I will summarize the bullet points of this section which cover the preliminary knowledge and the detailed procedure explanation of Multi-Agent Safety-Aware Apprenticeship Learning in the following parts:
\begin{enumerate}
\item (\ref{chapter4.1}) Detailed Definition of Markov Game 
\item (\ref{chapter4.2}) Multi-agent Reinforcement Learning
\item (\ref{chapter4.3}) Multi-agent Apprenticeship Learning via Inverse-Reinforcement Learning
\begin{enumerate}
\item (\ref{chapter4.3.1}) Learning from Expert Demonstration in Markov Game
\item (\ref{chapter4.3.2}) Multi-Agent Apprenticeship Learning via Inverse Reinforcement Learning Assumption.
\item (\ref{chapter4.3.3}) Optimal Policy Generation Algorithm Notation Demonstration by following Multi-agent Apprenticeship Learning Assumption.
\item (\ref{chapter4.3.4}) Optimal Policy Generation Algorithm Demonstration by following Multi-agent Apprenticeship Learning Assumption.
\end{enumerate}
\item (\ref{chapter4.4}) PCTL Model Checking in Multi-agent Apprenticeship Learning
\item (\ref{chapter4.5}) The Framework for Multi-agent Safety-Aware Apprenticeship Learning
\item (\ref{chapter4.6}) Counterexample-Guided Multi-Agent Safety-Aware Apprenticeship Learning
\item (\ref{chapter4.7}) Problem Solved by Multi-Agent Safety-Aware Apprenticeship Learning and the Extensions.
\end{enumerate}

\subsection{Detailed Definition of Markov Game}
\label{chapter4.1}
After understanding the details of the framework of single-agent reinforcement learning in Markov Decision Process(MDP) in the previous sections \ref{chapter2.2} and \ref{chapter2.3}, I will give detailed definition about what the Markov Game to help you understand the multi-agent reinforcement learning in the following section.

Recalling from the previous section \ref{chapter2.4} of the brief definition of Markov Game, we know that Basically, Markov Games are the generalization of the Markov Decision Processes(MDPs) to the case of $N$ interacting agents and a Markov Game is defined as a tuple $(S, \gamma, A, P, \triangle, r)$ via\cite{Hu1999}: 
\begin{enumerate}
\item A set of states $S$, which is the total global joint states of the agents with all state position possibilities
\item $N$ sets of actions $(A_{i})_{i->N}$;
\item The function $P$: $S$ $\times$ $A_{1}$ $\times$ $A_2$ $\times$ $...$ $\times$ $A_{N}$ $\longrightarrow$ $P(S)$ describes the stochastic transition process between states, where $P(S)$ means the set of probability distributions over the set $S$;
\item By giving that we are in state $s^{t}$ at time $t$, and the agent takes actions \{$a_{1}$ ,....., $a_{N}$\}, the state transitions to $s^{t+1}$ with probability $P(s^{t+1} | s^{t}, a_1,.......,a_N)$;
\item By taking the actions, each agent $i$ obtains a bounded reward given by a function $r_i$: $S$ $\times$ $A_{1}$ $\times$ $A_2$ $\times$ $...$ $\times$ $A_{N}$ $\longrightarrow$ $R$;
\item The function $\triangle$ $\in$ $P(S)$ specifies the probability distribution over state space $S$;
\item $\gamma$ $\in$  [0,1) is the discount factor which describes how future rewards attenuate when a sequence of transitions is made;
\end{enumerate}

To have a closer look at the Markov Game, let's consider the process that can be observed at discrete time point $t$ = 0, 1, 2, 3, . . .n. At each time point $t$, the state of the process is denoted by $s_{t}$. Assume $s_{t}$ takes values from the global joint state set $S$, the process is controlled by $N$ numbers of decision makers, referred to as agent 1, ..., $N$, respectively.

In state $s$, each agent independently chooses actions $a_1$ $\in$ $A_1$, $a_2$ $\in$ $A_2$, ..., $a_N$ $\in$ $A_N$ and receives the rewards $r_1$($s$, $a_1$, $a_2$, ...,$a_N$),  $r_2$($s$, $a_1$, $a_2$, ...,$a_N$), ...,  $r_N$($s$, $a_1$, $a_2$, ...,$a_N$). 

When  $r_1$($s$, $a_1$, $a_2$, ...,$a_N$) + $r_2$($s$, $a_1$, $a_2$, ...,$a_N$) + ... + $r_N$($s$, $a_1$, $a_2$, ...,$a_N$) = 0 for all $s$, $a_1$, ..., $a_N$, the game is called $zero$ $sum$. When the sum of reward functions is not restricted to 0 or any constant, the game is called a $general$ $sum$ game.

It's assumed that for every $s$ and $s'$ $\in$ $S$ , the transition from $s$ to $s'$ given that the players take actions $a_1$ $\in$ $A_1$, $a_2$ $\in$ $A_2$, ..., $a_N$ $\in$ $A_N$, is independent of time. So, this is saying that there exist stationary transition probability $p$($s'$ $|$ $s$,  $a_1$, $a_2$, ...,$a_N$) for all time point $t$ = 0, ..., $N$ satisfying as follow:
\begin{equation} \label{eq:sum_prob}
\sum_{s'=1}^{N} p(s' | s, a_1, a_2, ... , a_N) = 1
\end{equation}
The objective of each agent is to maximize the discounted sum of rewards. Let's assume we have a discount factor $\gamma$ $\in$ [0,1) and assume $\pi_1$, $\pi_2$, $\pi_3$, ..., $\pi_N$ are the policies of the agents respectively. Then, for a given initial joint state $s$, all players receive the following values from the Markov Game:

\begin{equation} \label{eq:reward value1}
V^{1}(s, \pi_1, \pi_2, ..., \pi_N) = \sum_{t=0}^{\infty}\gamma^{t}* E(r_{t}^{1} |\pi_1, \pi_2, ..., \pi_N, s_0 = s ) 
\end{equation}
\begin{equation} \label{eq:reward value2}
V^{2}(s, \pi_1, \pi_2, ..., \pi_N) = \sum_{t=0}^{\infty}\gamma^{t}* E(r_{t}^{2} |\pi_1, \pi_2, ..., \pi_N, s_0 = s ) 
\end{equation}
\begin{equation} \label{eq:reward valueN}
V^{N}(s, \pi_1, \pi_2, ..., \pi_N) = \sum_{t=0}^{\infty}\gamma^{t}* E(r_{t}^{N} |\pi_1, \pi_2, ..., \pi_N, s_0 = s ) 
\end{equation}
At here we define the policy $\pi$ as a set of all agents' individual policies where $\pi$ = ($\pi_0$, $\pi_1$, $\pi_2$, ..., $\pi_N$)  is defined over the entire process of the Markov Game. At time t, the policy, which can be called as the decision rule as well,  $\pi_t$ is defined for all agents. 

The decision rule $\pi$ is called stationary policy if and only if the decision rule is not changing regarding with the change of time $t$. That is saying all decision rule in the range of ($\pi_0$, ...., $\pi_t$) is fixed over the change of time $t$ and the decision rule $\pi$ is called the behavior policy if $\pi_t$ $=$ $f(h_{t})$, where $h_{t}$ is the history up to time $t$.
\begin{equation} \label{eq:history of agent policy}
h_{t} = (s_{0}, a_{0}^{1}, a_{0}^{2}, ..., a{0}^{N}, s_{1}, a_{1}^{1}, a_{1}^{2}, ..., a{1}^{N}, ..., s_{t-1}, a_{t-1}^{1}, a_{t-1}^{2}, ..., a_{t-1}^{N}, s_{t})
\end{equation}Based on the equation above, if the $h_{t}$ $=$ $\oslash$, then the  $\pi$ is stationary policy which is a special case of the behavior policy. 

The decision rule assigns mixed policies to different states. A decision rule of a stationary policy has the following form $\hat{\pi}$ $=$ ($\hat{\pi}(s^{1})$, $\hat{\pi}(s^{2})$, ...,  $\hat{\pi}(s^{N})$), where $N$ is the maximal number of states and  $\hat{\pi}$(s) is the mixed policies under state $s$. 

One of the key concepts in Markov Game is each agent in the game environment should reach a equilibrium, and many kinds of equilibrium have existed for the Markov Game. In our work, we are going to focus on one equilibrium implementation: Nash Equilibrium. The definition of Nash Equilibrium requires that: each agent's policy is the best response to the others' policy. If we assume all agents in Markov game follow the stationary policy, which mean s the decision rule $\pi$ is fixed over time t in the Markov game environment, we are assuming that there always is a Nash Equilibrium exist for any Markov game and the following theorem holds:

\textbf{Theorem 1}: \textit{Every general-sum discounted game possesses at least one equilibrium point in stationary policy}.

In Markov Game, a Nash Equilibrium point is a policy set such as ($\pi_{*}^{1}$, $\pi_{*}^{2}$, ...., $\pi_{*}^{N}$) which. The $Nash$ $Equilibrium$ for the Markov Game at all state $s$ $\in$ $S$ can be defined as:

\begin{equation} \label{Nash equilibrium 1}
V^{1}(s, \pi_{*}^{1}, \pi_{*}^{2}, ...., \pi_{*}^{N}) = V^{1}(s, \pi_{1}, \pi_{*}^{2}, ...., \pi_{*}^{N}) \ \forall\pi_{1} \in \Pi
\end{equation}
$and$
\begin{equation} \label{Nash equilibrium 2}
V^{1}(s, \pi_{*}^{1}, \pi_{*}^{2}, ...., \pi_{*}^{N}) = V^{1}(s, \pi_{*}^{1}, \pi_{2}, ...., \pi_{*}^{N}) \ \forall\pi_{2} \in \Pi
\end{equation}
$and$
\begin{equation} \label{Nash equilibrium 2}
V^{1}(s, \pi_{*}^{1}, \pi_{*}^{2}, ...., \pi_{*}^{N}) = V^{1}(s, \pi_{*}^{1}, \pi_{*}^{2}, ...., \pi_{N}) \ \forall\pi_{N} \in \Pi
\end{equation}
At the same time, in order to visualize the stages of the Markov game, we can view each stage of Markov game as a N-Matrix game in the figure 4.1.

\begin{figure}[h!]
	\centering
	\includegraphics[width=1.0\linewidth]{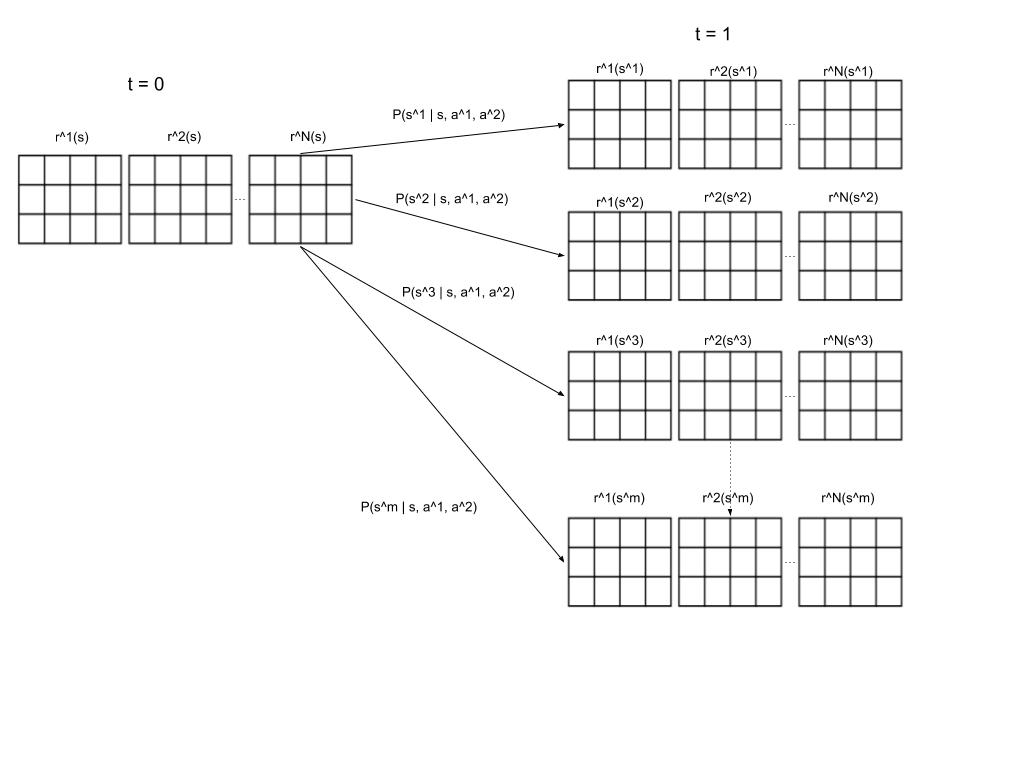}
	\caption{N-Matrix Game in Markov Game}
	\label{fig:N-Matrix Game}
\end{figure}

At each period of Markov Game, agents from agent 1 to agent N would take actions independently and receive their rewards according to the N-Matrix game ($r^{1}$, $r^{2}$, ..., $r^{N}$) under state s. 

\subsection{Multi-Agent Reinforcement Learning}
\label{chapter4.2}
After we understand the key concepts in single-agent reinforcement learning and Markov Game, in this subsection, I will explain the multi-agent reinforcement learning in detail. 

The main issue associating with the learning process in multi-agent environment is the state-action conjectures between multiple agents. It's true that the other agents could be treated as a part of the learning environment from the single agent perspective, but a problem arises when we are making symmetric decisions on according all of the agents' status in our model

So, at this moment, the Nash equilibrium concept characterizes some steady-state balance relationship among the agents in the environment and would help us with solving our issue mentioned above. In the multi-agent environment, all agents should be optimized at the same moment, and the Nash equilibrium, which is a consistent joint optimization among agents, represents the logical multi-agent extension from the single-agent optimization perspective. 

Nash equilibrium is a game's steady-state play, where each agent in the game holds the right expectation about the other agents' behaviors and should act rationally based on them. By saying All agents should act rationally based the other agents' behavior when they're following the Nash equilibrium in the environment, we are talking about each agent's policy is the best response to the other agents' policies assuming all agents have the common knowledge of rationality.\cite{Perolat2017}

When the agents do not have access to their own or other agents reward functions, we call this case as Markov Game with incomplete information, and Nash equilibrium cannot be applied due to the incomplete information of the game environment. \cite{Perolat2017}\cite{Gensbittell2012}

As a result, in this case, we'll assume that, at each time period, the agent would be able to observe the immediate reward of all other agents at each time period $t$. By following this way, the agents would gradually complete the missing information from the incomplete Markov game environment and build up the reward functions of all other agents. So, following the same logic, if the agents initially do not know their own or the other agents' transition probabilities in the environment, they can gradually learn them by playing the game repeatedly and construct the transition probability matrix finally. So, all missing information in the game should be completed and then we can apply apply Nash equilibrium to the game environment. \cite{Perolat2017}

Similar to single-agent reinforcement learning, in multi-agent reinforcement learning environment, we will also apply Q-Learning for the purpose of conducting multi-agent learning process. 

At here, multi-agent Q-learning mainly serves for two purposes:
\begin{enumerate}
\item Q-learning serves as a method which can computationally solves for Nash equilibrium by having no information with the transition probabilities in the multi-agent environment.
\item Under complete information game environment, Q-value generated from Q-learning process would provide the best approximation to the optimal values.
\end{enumerate}

So, after explaining the purpose of applying Q-learning in the multi-agent game environment, I will explain how Q-learning works within multi-agent learning environment in the following paragraph.

First, we need to extend and redefine the Q-function from single-agent learning environment to multi-agent learning environment. Recalling from previous subsection, the Q-function defined in the Single-Agent reinforcement learning environment. Since the game environment is extended to multi-agent case, assuming for a n-agent Markov game, we'll define a Q-value following the Nash equilibrium for agents $x$ and $x$ = 1, 2, 3, 4, .....$n$ as:

\begin{equation} \label{eq:Q-learning value single agent rewritten *}
Q_{*}^{x}(s, a^1, a^2, ..., a^n) = r^{x}(s, a^1, a^2, ..., a^n) + \gamma* \sum_{s' \in S}p(s'|s, a^1, a^2, ..., a^n)V(s',\pi_{*}^{1}, \pi_{*}^{2}, ..., \pi_{*}^{n}))
\end{equation}
Based on the equation above, the Q-value following the  Nash equilibrium is defined on state $s$ and joint action ($a^1$, $a^2$, ......., $a^n$) and it's the total discounted reward associating with the discount factor $\gamma$ received by the agent $x$ at the time that all the agents play the joint action ($a^1$, $a^2$, ......., $a^n$) at state $s$ and follow the joint policies ($\pi_{*}^{1}$, $\pi_{*}^{2}$, ..., $\pi_{*}^{n}$) satisfying the Nash equilibrium. 

In order to allow agent to learn the Q-values following Nash equilibrium, an agent needs to maintain $n$ Q-tables, for each agents in the game environment from 1 to $n$. For agent $x$, an element of the Q-table of itself $Q^{x}$ is represented by $Q^{x}(s, a^1, a^2, ..., a^n)$. If we assume there are $m$ number of states in total, $n$  number of agents in the environment, and use $|A^i|$ to represent the size of action space $A^i$, then the total number of entries of for a single agent would be $m\Pi_{1}^{n}|A^{i}|$. Since the total number of entries of the environment of an agent is $m\Pi_{1}^{n}|A^{i}|$ and we have $n$ agents joint with the agent $x$, then the total entries for the single agent needed to maintain is $nm\Pi_{1}^{n}|A^{i}|$ in the multi-agent game environment.

If we assume all action space are the same, in which $|A^1| = |A^2| = ... = |A^n|= A$, the space memory would be taken is $nm|A|^{n}$. Therefore, as the the number of agents increase, the size of the space memory would would explore. Due to the consideration of memory intake, it's important to present the action space $|A^i|$ compactly for the purpose of saving memories. 

\begin{figure}[h!]
	\centering
	\includegraphics[width=0.9\linewidth]{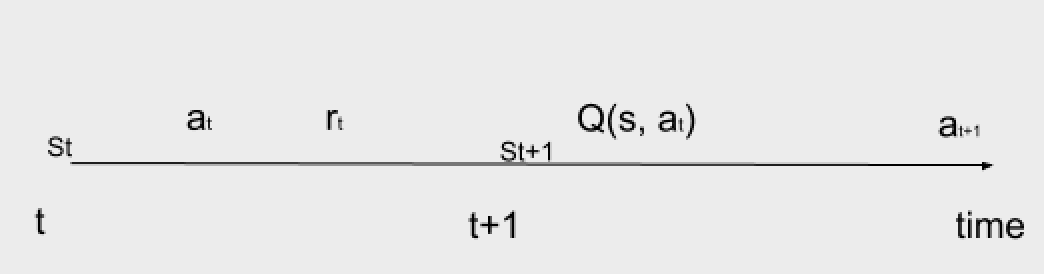}
	\caption{Single-agent Q-value Update through Time}
	\label{fig:single-agent-q}
\end{figure}

Similar to the single-agent Q-value updating process, in Markov game, the agents would also update their Q-values when all agents observe their current/next states $s/s'$, actions having been taken and rewards having been received. However, the different is the updating rules between the single-agent environment and the multi-agent environment. In multi-agent environment, we can't  update agents' Q-values only by considering maximizing the actions since because the Q-values are depending on the joint actions of all other agents. 

\begin{figure}[h!]
	\centering
	\includegraphics[width=0.9\linewidth]{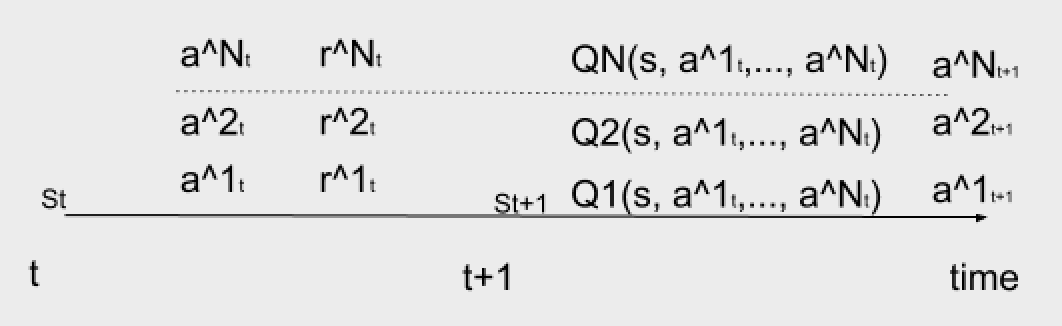}
	\caption{Multi-agent Q-value Update through Time}
	\label{fig:multi-agent-q}
\end{figure}

As a result, in order to correctly update the Q-values in the multi-agent environment for generating joint optimal policies, we followed the algorithm developed by Junling and Michael\cite{Hu1999} adapting the Nash Equilibrium to update the Q-values and generate the optimal policies in the multi-agent environment. The algorithm is defined below in detail.

Let $Q^{x}$ $=$ $(Q^{x}(s^1), ..., Q^{x}(s^m))$ be the agent $x$'s Q table and we have $m$ number of states in total. $Q^{x}(s^{i})$ is the Q-table under state $s^{i}$ and each element in the Q-table is defined as $Q^{x}(s^{i}, a^1, a^2, ..., a^n)$. Assuming there are $n$ agents exist in the Markov game environment, the total number of entries in the Q-table  $Q^{x}(s^{i})$ for agent $x$ at sate $s^i$ is $\Pi_{1}^{n}|A^{i}$. Agent $x$ updates the Q-values by following the equation below in the multi-agent environment:

\begin{equation} \label{eq:Q-learning value mul agent  *}\cite{Hu1999}
Q_{t+1}^{x}(s, a^1, a^2, ..., a^n) = (1-\alpha_{t})Q_{t}^{x}(s, a^1, a^2, ..., a^n) + \alpha_{t}[r_{t}^{k} + \gamma(\pi^1(s_{t+1}), ..., \pi^{n}(s_{t+1})))Q_{t}^{x}(s_{t+1})]
\end{equation}
where ($\pi^1(s_{t+1}), ..., \pi^{n}(s_{t+1})$) is the joint policy following the Nash equilibrium for the Markov game with complete information, which is assuming agent $x$ knows all Q-values of the other agents in the environment from $Q_{t}^{1}(s_{t+1})$to $Q_{t}^{n}(s_{t+1})$ for deriving all optimal policies.

If it's in the case of not having complete information in the Markov game environment and we don't know other agents' Q-values, then agent $x$ needs to learn all the missing information by itself by repeatedly playing the game. As agent $x$ plays the game, agent $x$ observes other agents' rewards and previous actions, which can be used to update agent $x$'s conjectures on other agents' Q-table. Then agent $x$ updates its belief on agent $y$'s Q-value, for all $x\neq y$, following the rule below:

\begin{equation} \label{eq:Q-learning value mul agent  *}
Q_{t+1}^{y}(s, a^1, a^2, ..., a^n) = (1-\alpha_{t})Q_{t}^{y}(s, a^1, a^2, ..., a^n) + \alpha_{t}[r_{t}^{k} + \gamma(\pi^1(s_{t+1}), ..., \pi^{n}(s_{t+1})))Q_{t}^{y}(s_{t+1})]
\end{equation}
and based on this rule, the Multi-Agent Q-learning Algorithm with Nash Equilibrium follows below.\cite{Hu1999}


\begin{algorithm}
\caption{Multi-Agent Q-learning Algorithm with Nash Equilibrium}
\begin{algorithmic}[1]
\STATE  \textbf{Initialize}
\STATE\hspace{\algorithmicindent}  $\forall$ $s$ $\in$ $S$, $\forall$ $a^x$ $\in$ $A^x$ , $x$ in $1, ...., n$, let $Q_{t}^{x}(s, a^1, a^2, ..., a^n)$ = 0.
\STATE\hspace{\algorithmicindent}  \textbf{Loop}:
\STATE\hspace{\algorithmicindent}\hspace{\algorithmicindent}  Choose action $a_t^i$
\STATE\hspace{\algorithmicindent}\hspace{\algorithmicindent}  Observe $(r_t^1, ...., r_t^n)$ ; $(a_t^1, ...., a_t^n)$, and $s_{t+1}$
\STATE\hspace{\algorithmicindent}\hspace{\algorithmicindent}  Update $Q^y$ for $y$ = $1, 2, ......., n$
\STATE\hspace{\algorithmicindent}\hspace{\algorithmicindent}  $Q_{t+1}^{y}(s, a^1, a^2, ..., a^n) = (1-\alpha_{t})Q_{t}^{y}(s, a^1, a^2, ..., a^n) + \alpha_{t}[r_{t}^{k} + \gamma(\pi^1(s_{t+1}), ..., \pi^{n}(s_{t+1})))Q_{t}^{y}(s_{t+1})]$, where  ($\pi^1(s_{t+1}), ..., \pi^{n}(s_{t+1})$) is the mixed policy following the Nash equilibrium for the Markov game with complete information, which is assuming agent $x$ knows all Q-values of the other agents in the environment from $Q_{t}^{1}(s_{t+1})$to $Q_{t}^{n}(s_{t+1})$ for deriving all optimal policies.
\STATE\hspace{\algorithmicindent}  \textbf{Let $t$ $+$=1}
\end{algorithmic}
\end{algorithm}


\subsection{Multi-agent Apprenticeship Learning via Inverse-Reinforcement Learning}
\label{chapter4.3}
After explaining the Multi-agent reinforcement learning and Markov game in detail in the previous subsections, which are the key prerequisite knowledge for implementing the Markov game extension to our project, we can finally look at the main part of our project, which is extending the single-agent apprenticeship learning to multi-agent environment. 

Recalling from section \ref{chapter2.1}, we have defined single-agent apprenticeship learning is a kind of learning from demonstration techniques where the reward function a Markov decision process is unknown to the learning agent, and the agent has to derive a good policy by observing an expert's demonstrations. Also, apprenticeship learning is mainly actualized via inverse-reinforcement learning. As a result, if we want to extend the single-agent apprenticeship learning to multi-agent apprenticeship learning, the first task we should finish is extending the inverse reinforcement learning from single-agent scenario to multi-agent scenario. 

\subsubsection{Learning from Expert Demonstration in Markov Game}
\label{chapter4.3.1}
According to Abbeel and Ng\cite{Ng2000}, in apprenticeship learning, because the reward function is unknown, the reward function is expressible as a linear combination of known state features. The expert demonstrates their task by maximizing the reward function and the agent tries to derive a policy that can match the feature expectations of the expert's demonstrations through inverse reinforcement learning.

In section \ref{chapter3.1}, we have specifically defined how the inverse reinforcement learning is implemented in single-agent apprenticeship learning. In single-agent inverse reinforcement learning, the main objective is to recover the reward function $R$ of the MDP \{$S$, $A$, $P$, $\gamma$, $s_0$, $R$\} from a set of $m$ trajectories $\tau$ = {$\tau_1$, $\tau_2$, ...., $\tau_m$} demonstrated by the expert, where each trajectory $\tau$ is defined as $\tau$ = $\{(s^t, a^t)\}_{t=1}^{T}$, where $t$ represents the iteration time that $t$ $\in$ $T$. 

In order to achieve single-agent apprenticeship learning via inverse reinforcement learning, we assume that the reward function of the MDP is a linear combination of known state features vector, such as $R(s)$ = $w^T$ $f(s)$. 
\begin{enumerate}
\item State features $f(s)$ $\implies$ $[0, 1]$ is a vector of known features over states $S$. 
\item $w$ $\in$ $R$ is an unknown weight vector that satisfies $||w_2||$ $\leq$ 1.
\end{enumerate}

However, in multi-agent inverse reinforcement learning, assuming we have $N$ agents in the game environment, the main objective is recovering the all agents' reward function $r_i$, where $r_i \in r$ and $r = (r_1, r_2,...,r_N)$ from the Markov Game  $(S, \gamma, A, P, \triangle, r)$ through $N$ corresponding experts' demonstrations.

In order to differentiate the expert demonstrated trajectories in single-agent learning environment and multi-agent learning environment, we define the joint expert demonstrated trajectories in multi-agent learning environment as $\hat{\tau}$. 

Because in the multi-agent learning environment, all agents' actions are jointed together, so the expert trajectory demonstration extension from single-agent to multi-agent environment will follow with:
\begin{enumerate}
\item $1$ x $m$ trajectories $\tau$ = {$\tau_1$, $\tau_2$, ...., $\tau_m$} $\longrightarrow$ $N$x$m$ joint trajectories $\hat{\tau}$ = {($\tau_{1}^{1}$, $\tau_{2}^{1}$, ..., $\tau_m^1$), ($\tau_{1}^{2}$, $\tau_{2}^{2}$, ..., $\tau_m^2$)...., ($\tau_1^N$, $\tau_2^N$,..., $\tau_m^N$)}.
\end{enumerate}

Because the learning environment has changed from MDP to Markov Game, all agents' status in the learning environment should have steady-state balance relationship and be optimized at the same moment based on the concept from the previous Multi-agent Reinforcement Learning section \ref{chapter4.2}. Due to the reason shows above and the reason that the we're having $N$ numbers of experts' demonstrations, which decide the agent's status in the learning environment, the experts demonstrations showing in the paragraph above should be set to follow the Nash Equilibrium in order to make the agents have steady-state balance relationship in the multi-agent learning environment.\cite{Inga2019} More relevant works would be seen in the next sections.

\subsubsection{Multi-Agent Apprenticeship Learning via Inverse Reinforcement Learning Assumption}
\label{chapter4.3.2}
At the same time, due to the reason of environment extension, the apprenticeship learning assumes the inverse reinforcement learning would be adjusted as well in the Markov game learning environment. Because we assume there exists $N$ agents in the learning environment, assuming the state feature is a matrix with size of $Q$x$Q$ in the single-agent environment, then it would change to a $N$-dimensional matrix with size of $Q_1^N$x$Q_2^N$x$Q_3^N$x....$Q_n^N$, where each agent would run on $Q^N$ numbers of joint states,including the possibility that agents arrive at the same states simultaneously in the multi-agent learning environment. The extension of state features should follow with:
\begin{enumerate}
\item If we have $N$ agents in the multi-agent learning environment and $Q^N$ states for each agent to go,
\item Then, the state feature matrix changes with: $Q$x$Q$ $\longrightarrow$ $Q_1^N$x$Q_2^N$x$Q_3^N$x....$Q_n^N$, where $n$ $\in$ $N$.
\end{enumerate}

Also, the unknown weight vector $w$, which has size of $Q$x1 in single agent case, would be extended to a matrix with size of $Q_1^N$x$Q_2^N$x$Q_3^N$x....$Q_n^N$ as well, following with:
\begin{enumerate}
\item If we have $N$ agents in the multi-agent learning environment and $Q^N$ states for each agent to go,
\item Then, the unknown weight vector changes following: $Q$x1 $\longrightarrow$ $Q_1^N$x$Q_2^N$x$Q_3^N$x....$Q_n^N$, where $n$ $\in$ $N$.
\end{enumerate}

So, the joint reward functions of Markov game would be a combination of state features matrix, such that $R(s)$ = $W$ $\times$ $f(s)$ where $R(s)$ is a $N$-dimensional joint reward matrix with size of $Q_1^N$x$Q_2^N$x$Q_3^N$x....$Q_n^N$used for generating the joint policy for all agents . 
\begin{enumerate}
\item $f(s)$ $\implies$ $[0, 1]$ is a $Q_1^N$x$Q_2^N$x$Q_3^N$x....$Q_n^N$ matrix of known features for all $N$ numbers of agents over joint states $S$, for $\forall$ $s$ $\in$ $S$. 
\item $W$ $\in$ $R$ is an $Q_1^N$x$Q_2^N$x$Q_3^N$x....$Q_n^N$ unknown weight matrix.
\end{enumerate}

\subsubsection{Optimal Policy Generation Algorithm Notation Demonstration by Following Multi-agent Apprenticeship Learning Assumption}
\label{chapter4.3.3}
So, in the previous section \ref{chapter3.1.2}, we have known, in order to extract the single-agent's optimal policy via inverse reinforcement learning, we need to estimate the expect features, which are the expected values of cumulative discounted state features $f(s)$ by following policy $\pi$ derived by the expert's demonstrated $m$ trajectories, such that $\mu_E$ = $E[\sum_{t=0}^{\infty}\gamma^t | \pi]$.\cite{ZhouLi2018}
\begin{enumerate}
\item $\mu_E$ denotes the expected features of the unknown expert's policy $\pi_E$.
\item $\gamma^t$ denotes the unknown weight vector satisfies $||\gamma_{2}^{t}||$ $\leq$ 1 at time $t$.
\item State features f(s) $\implies$ [0,1] is a vector of known features over states S.
\item $\mu_E$ can be approximated by the expected features of expert's $m$ demonstrated trajectories, such that:
\begin{itemize}
\item $\mu_E$ = 1/$m$ $\sum_{\tau \in \tau_E}$ $\sum_{t=0}^{\infty}\gamma^{t}f(s^t)$, if the set of $m$ expert's demonstration trajectories' size are big enough.
\end{itemize}
\end{enumerate}

In order to retrieve the optimal policy $\pi^*$ of a single agent using the expected feature $\mu_E$, we define a error bound $\epsilon$ and we can retrieve the optimal policy of the agent by using $\mu_E$, if $\mu_E$ satisfies relationship that:
\begin{enumerate}
\item $||\mu_E - \mu_{\pi^{*}}||_2$ $\leq$ $\epsilon$, where $\mu_{\pi^*}$ is the expected feature of the optimal policy $\pi^*$,
\item and the expected feature value of the optimal policy $\pi^*$ can be calculated by Monte Carlo Method, value iteration or linear Programming.
\end{enumerate}

By knowing the multi-agent AL assumption and following it, we can estimate the expected features $\mu_E$ for the decision rule $\hat{\pi}$, where $\hat{\pi}$ = ($\pi_0$, $\pi_1$, $\pi_2$, ..., $\pi_N$) representing the joint policies of $N$ agents in the Markov game. Then, the expect features $\mu_E$ in Markov game would change to:
\begin{enumerate}
\item $\mu_E$ is the expected feature of the unknown expert's decision rule  $\hat{\pi_E}$.
\item $\mu_E$ = $E[\sum_{t=0}^{\infty}\gamma^t | \hat{\pi}]$ = 1/($N$x$m$) $\sum_{i=0}^{N}\sum_{\tau_i \in \hat{\tau_E}}$ $\sum_{t=0}^{\infty}\gamma^{t}f(s^t)$
\item $\gamma^t$ denotes the unknown weight matrix at time $t$.
\item $f(s)$ $\implies$ [0,1] is a matrix of known features for all $N$ numbers of agents over joint states $S$, for $\forall$ $s$ $\in$ $S$.
\end{enumerate}

So by having the expected features $\mu_E$ for the unknown expert's decision rule $\hat{\pi_E}$, we follow the same logic in the single-agent apprenticeship learning by defining a error bound $\epsilon$ to retrieve optimal decision rule $\hat{\pi^*}$ if the expected features $\mu_E$ satisfies the relationship that the Euclidean norm of its expected state feature difference with the state feature of the optimal decision rule $\hat{\pi^*}$ less or equal to the error bound $\epsilon$. 


\subsubsection{Optimal Policy Generation Algorithm Demonstration by Following Multi-agent Apprenticeship Learning Assumption}
\label{chapter4.3.4}
After giving the demonstrations about the optimal policy generation algorithm notations, I will give the explanation on the optimal policy generation algorithm by following the multi-agent apprenticeship learning assumption.

Assuming in iteration $i$, we have found a set of candidate decision rules, the joint policies, $\hat{\Pi}$ = (
$\hat{\pi_0}$,$\hat{\pi_1}$, $\hat{\pi_2}$, .....) and the corresponding expected features \{$\mu_{\hat{\pi}}$ $|$ $\hat{\pi}$ $\in$ $\hat{\Pi}$\}, then we can apply the mini-max algorithm to adjust the value of the unknown weight matrix $W$ for retrieving the optimal decision rule $\hat{\pi}$. So, we would have:
\begin{equation} \label{eq: opti pol gen markov game *}
\delta = \max_{W} \min_{\pi \in \Pi} W \times (\hat{\mu_E} - \mu_{\pi})       
\end{equation}
In this equation, the $W$ is the unknown weight matrix which is used to find the corresponding optimal decision rule $\hat{\pi_i^*}$  and the expected features of current expert's decision rule $\mu_{\hat{\pi_i}}$. The optimal Policy generation algorithm follows:
\begin{enumerate}
\item If the value of $\delta$ < $\epsilon$ (the predefined error bound):
\begin{enumerate}
\item The algorithm terminates and the decision rule $\hat{\pi_i}$ is produced as the optimal decision rule.
\end{enumerate}
\item else:
\begin{enumerate}
\item The algorithm goes on. The expected feature value from the current expert's decision rule $\mu_{\hat{\pi_i}}$ is added to the set of features and the decision rule $\hat{\pi_i}$ is added to the candidate decision rule set $\hat{\Pi}$ until the optimal decision rule  $\hat{\pi_i}$ is found.
\end{enumerate}
\end{enumerate}

\subsection{PCTL Model Checking in Multi-agent Apprenticeship Learning}
\label{chapter4.4}
In the previous section \ref{chapter2.5} and \ref{chapter3.2}, we have give explanation about how the PCTL model checking works for the single-agent safety-aware Apprenticeship learning. The PCTL model checking mainly serves as the logical model $verifier$ for verifying whether the generated policy by experts demonstrated $m$  trajectories satisfies the safety requirement. In our project, we assume that we are playing $N$-agent discrete-time Markov game. So, the model that we are applying with the PCTL is the Discrete-Time Markov Chain(DTMC).

Based on the Zhou and Li's algorithm for model checking\cite{ZhouLi2018}, in the single-agent safety-aware apprenticeship learning, they define the $pref(\tau)$ as the set of all $prefixes$  of trajectory $\tau$ including $\tau$ itself, then $\tau$ $\models_{min}$ $\Psi$ ( $\models_{min}$ means there is minimal satisfaction relationship between the trajectory $\tau$ and path formula $\Psi$ , which represents the property of a trajectory $\tau$ exists) iff:

\begin{equation} \label{minimal satisfaction}
(\tau \models \Psi) \land (\forall \tau^{'} \in  pref(\tau) \setminus \tau , \tau^{'} \nvDash \Psi)
\end{equation}
\begin{itemize}
\item In a easier way to explain this satisfaction relationship, we can utilize an example as such:
\begin{enumerate}
\item if $\psi$ $=$ $\Phi_1$ $\cup^{\leq k}$ $\Phi_2$, then for any finite trajectory, we have the minimal satisfaction relationship exists that:
\begin{itemize}
\item $\tau$ $\models_{min}$  $\Phi_1$ $\cup^{\leq k}$ $\Phi_2$, and only the final state in $\tau$ satisfies $\Phi_2$.
\end{itemize}
\end{enumerate}
\end{itemize}

In multi-agent safety-aware apprenticeship learning, we assume the joint expert demonstrated trajectories $\hat{\tau}$s used for generating the optimal decision rules $\hat{\pi}$ will follow the same minimal satisfaction relationship between the $\hat{\tau}$ and the path formula $\Psi$  such as:
\begin{equation} \label{multi-agent minimal satisfaction}
(\tau_i \models \Psi) \land (\forall \tau^{'} \in  pref(\tau \setminus \tau , \tau^{'} \nvDash \Psi) for  \tau_i \in \hat{\tau}
\end{equation}
As a result, let $P(\hat{\tau})$ be the Probability of transitioning along the joint trajectory $\hat{\tau}$ in the multi-agent learning environment and let $\tau_{\Psi}$ be the set of all finite trajectories that satisfies $\hat{\tau}$ $\models_{min}$ $\Psi$ , then the value of PCTL property $\Psi$ is defined as $P(\hat{\tau})$ =  $\sum_{i=0}^{N}$ $\sum_{\tau_i \in \hat{\tau_\Psi^i}}$ $P(\tau_i)$.

\begin{itemize}
\item So, in Markov game, by applying the Discrete-time-Markov-Chain (DTMC) $M_{\hat{\pi}}$ and a state formula $\Phi$, we have:
\begin{enumerate}
\item A \textit{counterexample} of $\Phi$ is a set $cex$ $\subseteq$ $\hat{\tau_E}$ that satisfies $\sum_{\hat{\tau} \in cex}$ $P(\hat{\tau})$ $>$ $p^{*}$, where $p^*$ is the probability of reaching the unsafe states in the learning environment.
\item $P(\hat{\tau})$ = $\sum_{\hat{\tau^{*}} \in \hat{\tau}}$ $P(\hat{\tau^{*}})$ is the sum of probability of all trajectories in joint trajectory set $\hat{\tau}$.
\item $CEX_{\Phi}$ $\subseteq$ $2^{\hat{\tau_{\Psi}}}$ is the set of all \textit{counterexamples} for a state formula $\Phi$.
\end{enumerate}
\end{itemize}

Once we defined our counterexamples $CEX_{\Phi}$ above, we can convert the DTMC $M_{\hat{\pi}}$ into a weighted directed graph. Then, we can use the converted DTMC to generate our counterexamples by solving the k-shortest paths problems or a hop-constrained k-shortest paths problems by following the algorithm in the section \ref{chapter2.6.2}.

\subsection{The Framework for Multi-agent Safety-Aware Apprenticeship Learning}
\label{chapter4.5}
From previous section \ref{chapter3.3} the framework for single-agent safety-aware apprenticeship learning, we have had a clear structure about the entire framework about how the safety-aware apprenticeship learning works in the single-agent learning environment. 

In the framework for multi-agent safety-aware apprenticeship learning, the logic is very similar with that of the framework of single-agent safety-aware apprenticeship learning but with several modifications. The framework of multi-agent safety-aware apprenticeship learning can be generalized to four parts:
\begin{enumerate}
\item The PCTL model checker verifies whether the current decision rule $\hat{\pi^*}$ \cite{Moldovan2012}, where $\hat{\pi^*}$ = ($\pi_0$, $\pi_1$,$\pi_2$,$\pi_3$ , ....., $\pi_n$), satisfies the state formula $\Phi$.
\item If candidate decision rule $\hat{\pi^*}$ satisfies the state formula $\Phi$, then we check whether our learning objective is met or not. Remember from the previous section \ref{chapter4.3.4} Optimal policy generation algorithm demonstration by following multi-agent apprenticeship learning algorithm, our learning objective is finding the optimal decision rule $\hat{\pi^*}$ by following the optimal policy generation algorithm in multi-agent apprenticeship learning defined previously:
\begin{enumerate}
\item If our learning objective is met, then we find the optimal decision rule . 
\item Otherwise, we add the checked policy to our candidate decision rule set $\hat{\Phi}$ and keep searching the decision rules which can satisfy our learning objective.
\end{enumerate}
\item If the candidate decision rule $\hat{\pi^*}$ doesn't satisfy the state formula $\Phi$, then we generate the corresponding counterexample denoted as $cex$ and continue the learning iteration. \cite{Jansen2012}
\item The learning iteration will not terminate unless the optimal decision rule $\hat{\pi^*}$ is found.
\end{enumerate}

The visualization of the framework for Multi-Agent Safety-Aware Apprenticeship Learning is shown in the figure 4.4.
\begin{figure}[h!] \label{fig: 4.4}
	\centering
	\includegraphics[width=1.0\columnwidth, scale= 1.0]{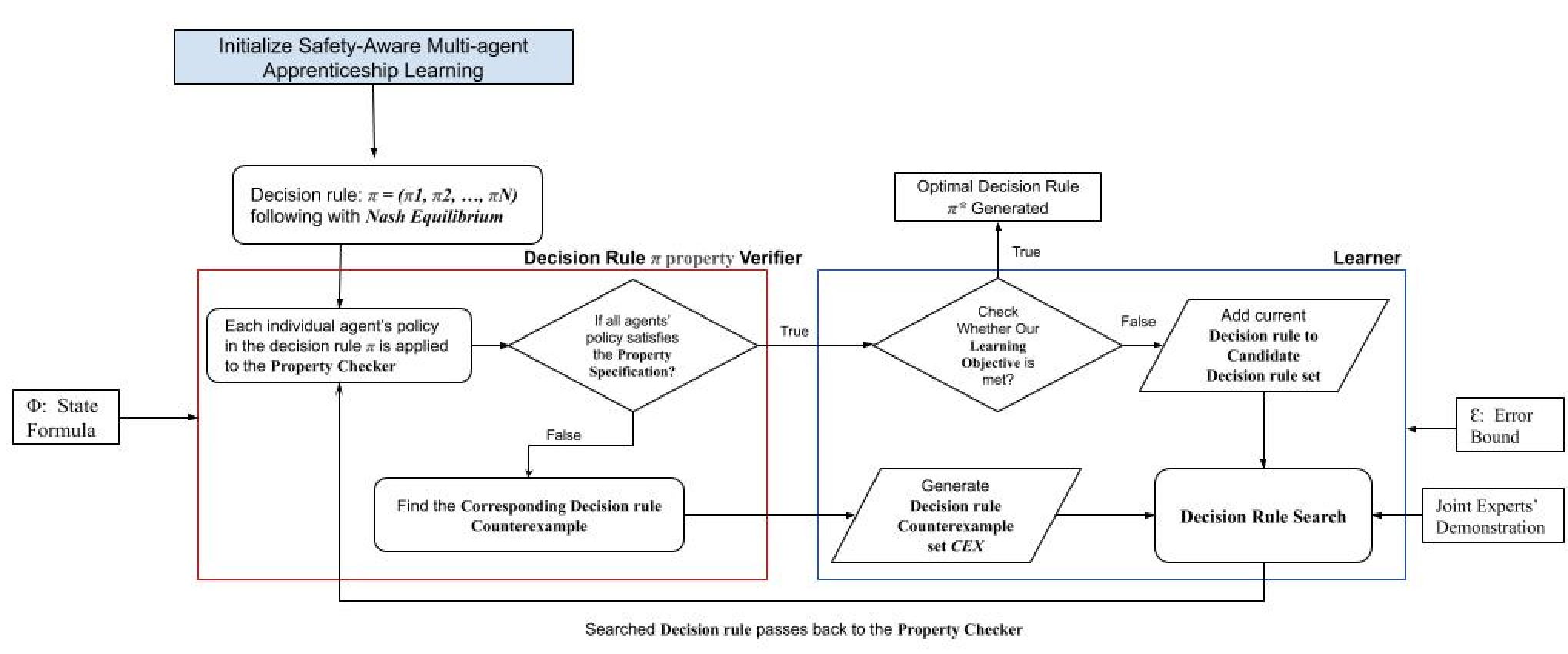}
	\caption{Multi-Agent Safety-Aware Apprenticeship Learning Framework}
	\label{fig:multi-agent-framework}
\end{figure}

After discussing the basic framework for the Multi-Agent Safety-aware Apprenticeship Learning, we can dive into more details about how this framework works.

By looking at the claims from Zhou and Li\cite{ZhouLi2018}, the Multi-agent AL algorithm can be finding a weight matrix $W$ under the condition that the expected reward generated from $\hat{\pi_E}$ maximally outperforms any mixture of the decision rule in the candidate decision rule set $\hat{\Pi}$.  

As a result, we can set the product of the weight matrix $W$ and the difference between the expected feature of the unknown experts' demonstration and the state feature of the candidate decision rule as 0 such that $W \times$ $(\mu_{\hat{\pi}} - \mu_{E})$ $=$  $0$, which has the maximal distance to the convex hall of the set \{$\mu_{\hat{\pi}}$ $|$ $\hat{\pi}$ $\in$ $\hat{\Pi}$\}. \cite{ZhouLi2018}\cite{Inga2019}, we can show that:
\begin{enumerate}
\item $W\times$ $\mu_{\hat{\pi}}$ $\geq$ $W\times$ $\mu_{\hat{\pi^{i}}}$ for all found decision rule $\hat{\pi^{i}}$ $\in$ $\hat{\Pi}$.
\item By performing the this kind of max-margin separation technique\cite{Ratlif2006}\cite{Huynh2009}, we can move the candidate decision rule's expected feature $\mu_{\hat{\pi}}$ closer to the expected feature demonstrated by expert $\mu_E$.
\end{enumerate}

In order to make max-margin separation technique more clear to understand, we will give a brief introduction about it here: 

\begin{enumerate}
    \item Max-margin methods are a competing approach to discriminating training that are well-founded in computational learning theory and have demonstrated empirical success in many applications.\cite{N.2000} They also have the advantage that they can be adapted to maximize a variety of performance metrics in addition to classification accuracy.\cite{Joachims2005} Max-margin methods have been successfully applied to structured prediction problems, such as in Max-Margin Markov Networks (M3Ns) and structural Support Vector Machines.\cite{Taskar2003} 
    \item In our work, we want to use the Maximum-Margin technique to learn such behaviors which are sequential, goal-directed structured over a space of policies in Markov Game.\cite{Ratlif2006}
\end{enumerate}

Therefore, we can similarly apply this technique, the max-margin separation technique, to maximize the distance between the candidate decision rule $\hat{\pi}$ and the decision rule counterexamples.\cite{Inga2019}\cite{ZhouLi2018}
\begin{enumerate}
\item Let $CEX$ = \{$cex_0$, $cex_1$ , $cex_2$ , ...\} denotes the set of counterexamples of the decision rules that do not satisfy the Specification $\Phi$ in the framework.
\item Maximizing the distance between the convex hulls of the set \{$\mu_{cex}$ $|$ $cex$ $\in$ $CEX$ \} and the set \{$\mu_{\hat{\pi}} $ $|$ $\hat{\pi}$ $\in$ $\hat{\Pi}$\} is equivalent to maximizing the distance between the parallel supporting hyperplanes of the two convex hulls in the Euclidean space.
\end{enumerate}

This is the formula for generating the counterexamples:
\begin{equation} \label{max margin separation}
\delta = \max_{W} \min_{\hat{\pi} \in \hat{\Pi}, cex \in CEX} W\times(\mu_{\hat{\pi}} - \mu_{cex})    
\end{equation}
In order to attain good performance similar to that of the expert, we still want to learn from $\mu_{E}$. Thus, the overall problem can be formulated as a multi-objective optimization problem, and formulate to the formula show below:
\begin{equation} \label{eq: Overall formula}
\max_{W} \min_{\hat{\pi} \in \hat{\Pi}, cex \in CEX, \tilde{\pi} \in \hat{\Pi}} (W\times(\mu_{\hat{\pi}} - \mu_{cex}), W\times(\mu_{\tilde{\pi}} - \mu_{cex}))     
\end{equation}

\subsection{Counterexample-Guided Multi-Agent Safety-Aware Apprenticeship Learning }
\label{chapter4.6}
Finally, we come to the final stage of the Multi-Agent Safety-Aware Apprenticeship Learning, which is the Counterexample-Guided Multi-Agent Apprenticeship Learning algorithm used for solve Multi-Agent SafeAL problem. We can regard this algorithm as a special case of the framework of the Multi-Agent Safety-Aware Apprenticeship Learning in the section \ref{chapter4.5} shown above. 

The one special add-on in this case to the original framework is using adaptive weighting scheme to weight the scheme from expected state feature of the expert demonstration $\mu_{E}$ with the separation from state feature of the decision rule counterexamples $\mu_{cex}$.

Originally, our framework works as the formula \ref{eq: Overall formula}. Now, after having the add-on to our framework, our framework works as:
\begin{multline} \label{eq: new framework}
\max_{W} \min_{\hat{\pi} \in \hat{\Pi_S}, \tilde{\pi} \in \hat{\Pi_S}, cex \in CEX} (k(\mu_{E} - \mu_{\hat{\pi}}) +  (1-k)(\mu_{\tilde{\pi}} - \mu_{cex})) , k \in [0,1] \\
W\times(\mu_E - \mu_{\hat{\pi}}) \leq W\times(\mu_E - \mu_{\hat{\pi'}}), \forall \hat{\pi'} \in \hat{\Pi_{S}} \\
W\times(\mu_{\tilde{\pi}} - \mu_{cex}) \leq W\times(\mu_{\tilde{\pi'}} - \mu_{cex^{'}}), \forall \tilde{\pi'} \in \hat{\Pi_{S}}, \forall cex^{'} \in CEX \\
\end{multline}

\begin{itemize}
\item The $K$ and $(1-K)$ are our weighting scheme as an add-on to the original framework.
\item Assuming $\hat{\Pi_{S}}$ $=$ \{${\hat{\pi_{1}}, \hat{\pi_2}, ...}$\} is a set of candidate decision rules, in which each individual agent's policy  satisfies the Specification $\Phi$.
\item Assuming $CEX$ = \{$cex_1$, $cex_2$ , $...$\} is a set of decision rule counterexamples.
\item We introduce a parameter $K$ into the formula 4.16 and change it into a weighted sum of optimization problem shown in the formula \ref{eq: new framework}.
\item It's important to note: the decision rule $\hat{\pi}$ and $\tilde{\pi}$ are different:
\begin{enumerate}
\item The optimal weight matrix $W$ can be use for generating the new decision rule $\pi_{W}$ by iterating our decision rule $\hat{\pi}$.
\item Then, we apply the PCTL model checker to see if $\pi_{W}$ satisfies $\Phi$:
\begin{enumerate}
\item Satisfy: We add the newly generate $\pi_{W}$ to the candidate policy set $\hat{\Pi_{s}}$.
\item Not satisfy: We generate a counterexample $cex_{\pi_{W}}$ and add it to the decision rule counterexample set $CEX$.
\end{enumerate}
\end{enumerate}
\end{itemize}

Detailed Counterexample-Guided Multi-Agent Safety-Aware Apprenticeship Learning Algorithm pseudo-code is divided into 2 parts and shown in page 45 and page 46.


\begin{algorithm} \label{algorithm: 2}
\caption{Counterexample-Guided Multi-Agent Safety-Aware Apprenticeship Learning Algorithm Part1}
\begin{algorithmic}[1]
    \STATE \textbf{Input}:
    \STATE {$\Phi$ $\gets$ Property Specification; $\epsilon$ $\longleftarrow$ Error bound for expected features, which is the learning objective.}
    \STATE {$\mu_E$ $\gets$ the expected feature of the unknown expert's demonstration joint trajectories $\hat{\tau}$ = {($\tau_{1}^{1}$, $\tau_{1}^{2}$, ..., $\tau_1^N$), ($\tau_{2}^{1}$, $\tau_{2}^{2}$, ..., $\tau_2^N$)...., ($\tau_m^1$, $\tau_m^2$,..., $\tau_m^N$)} following Nash Equilibrium.}
    \STATE {$M$ $\gets$ $A = (S \times A_1 \times A_2 \times ...\times A_N)$, partially known as Markov Game.}
    \STATE {$f(s)$ $\gets$ $A = (S \times A_1 \times A_2 \times ...\times A_N)$, known as a matrix of known features for all $N$ numbers of agents over joint states $S$, for $\forall$ $s$ $\in$ $S$.}
    \STATE {$\alpha$, $\sigma$  $\in$  (0,1)$\gets$ $\sigma$ is the error bound and $\alpha$ is the step length used for updating the adaptive weight scheme parameter $k$.}
    \STATE
    \STATE \textbf{Algorithm initialization}:
        \IF{$||\mu_E - \mu_{\hat{\pi_0}}||_E$ $\leq$ $\epsilon$}  
            \STATE return $\hat{\pi_0}$, where $\hat{\pi_0}$ is the initial \textbf{safe decision rule}.
        \ENDIF
        \STATE $CEX$ $\gets$ $\{\}$ , $\hat{\Pi_S}$ $\gets$ $\{\hat{\pi_0}\}$ , Initializing \textbf{(i) the decision rule counterexample set }$CEX$ and \textbf{(ii) the candidate decision rule set} $\hat{\Pi_S}$.
        \STATE $inf$ $\gets$ $0$, $sup$ $\gets$ $1$, $k$ $\gets$ $sup$, Initializing optimization for the weight scheme parameter $k$.
        \STATE $\hat{\pi_i}$ $\gets$ Decision rule learnt from the expected state feature of unknown expert Demonstration $\mu_E$.
\end{algorithmic}
\end{algorithm}

\begin{algorithm} \label{algo: 3}
\caption{Counterexample-Guided Multi-Agent Safety-Aware Apprenticeship Learning Algorithm Part2}
\begin{algorithmic}[1]
    \STATE Continue with \textbf{Algorithm 2}
    \STATE
    \STATE \textbf{Iteration} $i(i \geq 1)$ :\\
    \STATE
    \STATE \textbf{Decision Rule Property Verifier}:
        \STATE status $\gets$  PCTL-Model-Checker(M, $\hat{\pi_i}$, $\Phi$) .
        \IF{status = $Satisfy$}
            \STATE {we go to the learner section}.
        \ENDIF
        \IF{status = $Unsatisfy$}
            \STATE $cex_{\hat{\pi_i}}$ $\gets$ Decision-Rule-Counterexample-Generator(M, $\hat{\pi_i}$, $\Phi$).
            \STATE Add $cex_{\hat{\pi_i}}$ to $CEX$ and solve the state feature for $cex_{\hat{\pi_i}}$  and get corresponding state feature $\mu_{cex_{\hat{\pi_i}}}$. \textbf{Then, we go to the Learner}.
        \ENDIF
    \STATE
    \STATE \textbf{Learner}:
        \IF{status = $Satisfy$}
            \IF{$||\mu_E - \mu_{\hat{\pi_i}}||_E$ $\leq$ $\epsilon$}
            \STATE return the optimal decision rule $\hat{\pi*}$ $\gets$ $\hat{\pi_i}$ . At the same time, we terminate the learner here, since $\hat{\pi_i}$ is $\epsilon - close$ to $\hat{\pi_E}$.
            \ENDIF
            \STATE Add $\hat{\pi_i}$ to $\Pi_S$ , $inf$ $\gets$ $k$, $k$ $\gets$ $sup$, and Update $\Pi_S$, $inf$, and reset $k$.
        \ENDIF
            
        \IF{status = $Unsatisfy$}
            \IF{$|k - inf|$ $\leq$ $\sigma$}
            \STATE return the optimal decision rule $\hat{\pi*}$ $\gets$ $argmin_{\hat{\pi} \in \hat{\Pi_S}}$ $||\mu_E -\mu_{\hat{\pi}}||_E$. We terminate the learner because  $k$ is too close to its lower bound $inf$.
            \ENDIF
            
            \STATE $k$ $\gets$ $\alpha$ $\cdot$ $inf$ + $(1-\alpha) \cdot k$, Update our adaptive weight schema $k$ here.
        \ENDIF
        \STATE $W_{i+1}$ $\gets$ $argmax_W\min_{\hat{\pi} \in \hat{\Pi_S}, \tilde{\pi} \in \hat{\Pi_S}, cex \in CEX} W \times (k(\mu_{E} - \mu_{\hat{\pi}}) +  (1-k)(\mu_{\tilde{\pi}} - \mu_{cex}))$, Update our weight matrix here.
        \STATE $\hat{\pi_{i+1}}$, $\mu_{\hat{\pi_{i+1}}}$ $\gets$ Compute the optimal decision rule $\hat{\pi_{i+1}}$ and its expected features $\mu_{\hat{\pi_{i+1}}}$ for the Markov Game $M$ with reward $R$ = $W \times f(s)$.
    \STATE  
    \STATE Go to the next learning iteration, $i = i + 1$\
\end{algorithmic}
\end{algorithm}

\begin{itemize}
\item In this algorithm, $sup$ = 1, which is a constant.
\item $inf$, is a variable, and $inf$ $\in$ $[0, sup]$ for the upper and lower bound respectively.
\item The learner determines the value of $k$ within the bound $[inf, sup]$ in each decision rule search iteration depending on the outcome of the decision rule property verifier and use $k$ to solve the line 26 in the algorithm part2 pseudo-code.
\end{itemize}

Based on this algorithm, we can produce a general theorem showing:
\begin{enumerate}
\item Given the initial decision rule $\hat{\pi_{0}}$ , in which each individual agent's policy satisfies the property specification $\Phi$, this Counterexample-Guided Apprenticeship Learning Algorithm promises:
\begin{enumerate}
\item Producing a decision rule $\hat{\pi^{*}}$: such $\hat{\pi^{*}}$, in which each individual agent's policy  satisfies the property specification $\Phi$.
\item and such $\hat{\pi^{*}}$ has the performance is at least as good as that of the initial decision rule $\hat{\pi_{0}}$ when compare with the decision rule derived from expert demonstrations $\hat{\pi_{E}}$.
\end{enumerate}
\end{enumerate}

\subsection{Problem Solved by Multi-Agent Safety-Aware Apprenticeship Learning and the Extensions}
\label{chapter4.7}

The key extension that we made was extending the game environment $M$ from Markov Decision Process (MDP) to Markov Game (MG) in the Multi-Agent Safety-Aware Apprenticeship Learning comparing with the Single-Agent case, and the detailed differences are:
\begin{enumerate}
\item The $m$  expert demonstrated trajectories changed to $N\times m$ joint trajectories used for $N$ agent case.
\item The $N \times m$ expert demonstrated trajectories are forced to follow the Nash Equilibrium.
\item The weight vector $w$ used to calculate the reward $R$ in IRL process changed to weight matrix $W$ used for $N$ agent case.
\item The state features $s$ used to calculate the reward $R$ in IRL process changed from single state feature matrix to a joint state feature matrix used for the $N$ agent case. 
\item The  list of actions for the single agent become a list of joint actions in the $N$ agent case.
\item The Single-Agent Policy $\pi$ derived from expert demonstration becomes a decision rule $\hat{\pi}$ which contains joint agent policies for $N$ agent case.
\item PCTL model checker checks whether $N$ numbers of individual agent's policy in the decision rule $\hat{\pi}$ satisfy the property specification $\Phi$, instead of checking single policy each time in the Single-Agent case.
\item Counterexample $CEX$ contains the decision rule counterexamples, in which each contains $N$ numbers of policy counterexample for each individual agent in the $N$ agent case.
\end{enumerate}

In the Multi-Agent Safety-Aware Apprenticeship Learning, we solved the safety issue, the problem of the multiple agents having probability of reaching the unsafe states, by forcing the policies derived from joint demonstrated trajectories from the expert following the PCTL model checking specification $\Phi$.

The reason that agents are possible to reach the unsafe states in the Apprenticeship Learning is explained in the section \ref{chapter3.4}.


\newpage
\section{Experiment}
\label{chapter5}

In this section, I will give illustration about our project implementation and the experiment result. The contents of this section are divided to such subsections:
\begin{enumerate}
\item[(\ref{chaper5.1})] Problem Recap
\item[(\ref{chapter5.2})] Experiment Overview
\item[(\ref{chapter5.3})] Grid World Environment Example Experiment Evaluation in 2-Agent Scenario
\item[(\ref{chapter5.4})] Scalability Evaluation
\end{enumerate}

\subsection{Problem Recap}
\label{chaper5.1}
Let's assume there are some unsafe states in an Markov Game $M$ = $(S, \gamma, A, P, \triangle, r)$ [For symbol specification, please refer back to the section \ref{chapter4.1}]. A safety issue in Multi-Agent Apprenticeship Learning means that agents following the learnt decision rule $\hat{\pi}$ from the joint expert's demonstrations would have higher probability of entering those unsafe states than it should. 

There are multiple reasons can cause such problem:
\begin{enumerate}
\item  It is possible that the expert decision rule $\hat{\pi_E}$ has a high probability of reaching the unsafe states.
\item  Human experts often tend to perform only successful demonstrations that do not consider the negative conditions . In the training process, the lack of considering negative conditions will give rise to the problem that the learning agents reach the unsafe states, since they don't have awareness of reach those states.
\end{enumerate}

In order to solve such problem, we applied Counterexample-Guided Multi-Agent Apprenticeship Learning Algorithm, which is discussed in section \ref{chapter4.6} in detail.

\subsection{Experiment Overview}
\label{chapter5.2}
In our experiment, we evaluated our extended algorithm in section \ref{chapter4.6} in the case study environment: Grid World. The experiment was executed on Quad-core Intel i7-9750H processor running on 2.6GHz with memory 16GB. The tool for building our training environment is Python(2.7). 

The parameters for running Multi-Agent training process are set to:
\begin{enumerate}
\item  $\gamma = 0.99$: the learning rate, referred back to section \ref{chapter4.1},
\item $\epsilon = 10$: the learning objective error calculated between the state feature calculated from expert's trajectory demonstration and the state feature of the decision rule in the current iteration referred back to section \ref{chapter4.6} algorithm2,
\item $\sigma = 10^{-5}$: the error bound used for updating the adaptive weight scheme parameter $k$ referred back to section \ref{chapter4.6} algorithm2,
\item $\alpha = 0.5$ : the step length used for updating the adaptive weight scheme parameter $k$ referred back to section \ref{chapter4.6} algorithm2.
\item maximal training iterations are set to 200 referred back to section \ref{chapter4.6} algorithm3.
\end{enumerate}

\subsection{Grid World Environment Evaluation in 2-Agent Scenario}
\label{chapter5.3}
We use the $8\times8$, 2-agents Grid World environment to do the experiment demonstration and we assume agents can take actions independently and have no interactions with each other for constraining the complexity of our work.

\begin{figure}[h!]
	\centering
	\includegraphics[width=0.9\linewidth]{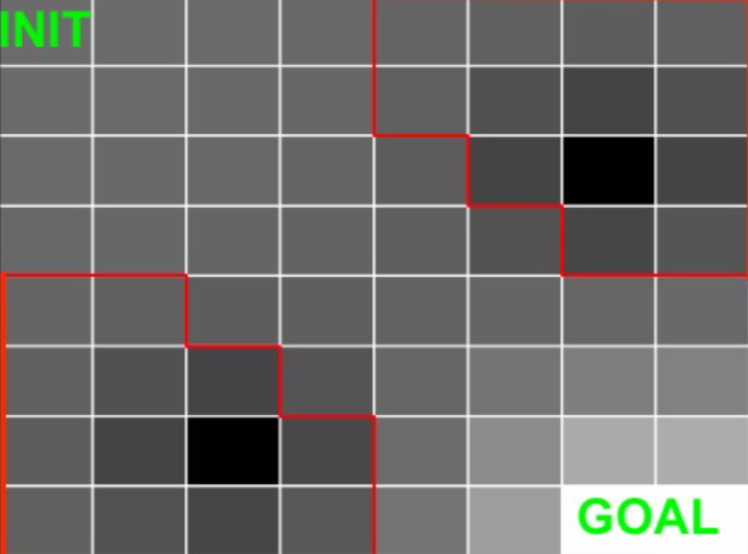}
	\caption{Initial reward mapping in $8\times8$ Grid World Environment }
	\label{fig:initial grid environment}
	
\end{figure}
\label{figure5.1}

In figure 5.1, we have the initial $8\times8$ Grid World Environment and all of the cell states with different colors represent the reward mapping to each cell state. In the environment, the agents are set to start at the upper-left cell state which is marked as INIT in green color and the goal for the agents is moving to the lower-right target cell states which are marked as GOAL in green color by taking many moving steps. The darker cell states in the environment have lower rewards than those of the lighter cell states. The two darkest cell states have the lowest rewards and the two white cell states have the highest rewards in the environment. The cell states which are surrounded by the red lines are set to be the unsafe area in the environment, and the agents should avoid from entering the unsafe areas.

In the Grid World environment, agents can jointly take 5 actions $(0: stay, 1: left, 2: down, 3: right, 4: up)$, which allow the agents to stay or move to the adjacent cell states with random stochastic probability. Since we have 5 actions for each individual agents, if we have $N$ agents in the environment, then the number of the joint actions would be $5^N$, such that ($(a_1, a_2, ...a_N)_1$, $(a_1, a_2, ...a_N)_2$, ..., $(a_1, a_2, ...a_N)_{5^N}$). For more details, please refer back to section \ref{chapter4.2}.

At the same time, if we assume each individual agent can move in a $Q\times Q$ grid matrix, in the Single-Agent learning Grid World environment, then when we have $N$ numbers of agents, there would be $Q^N \times Q^N \times...\times Q^N$ grid matrix for $N$ agents to move in the Multi-agent learning Grid Word environment. For more details, please refer back to the section \ref{chapter4.3.2}.

In our work, we set taking the actions $0:stay$, $1: left$ and $3: right$ to deterministic for each agent. This means, for each agent, if it takes action $0:stay$, $1: left$ or $3: right$, the probability of staying at the current cell state or moving to the corresponding adjacent cell state is always 1. So, the only moves that would be stochastic for the agents are taking action $2:down$ or $4:up$ with 0.5 probability.

In our example, because we have 2 agents and run the training in a $8\times8$ Grid World environment, then we inherently would have $8^2\times8^2$ joint grid matrix and $5^2$ possible joint actions for the 2 agents to move. We assume for the joint cell state $s$ that the agents are currently locating in, where $s$ $\in$ $S$, there is a corresponding the feature matrix exists calculated by $f(s)$, which is explained in the section \ref{chapter4.3.2}. 

\begin{figure}[h!]
	\centering
	\includegraphics[width=0.9\linewidth]{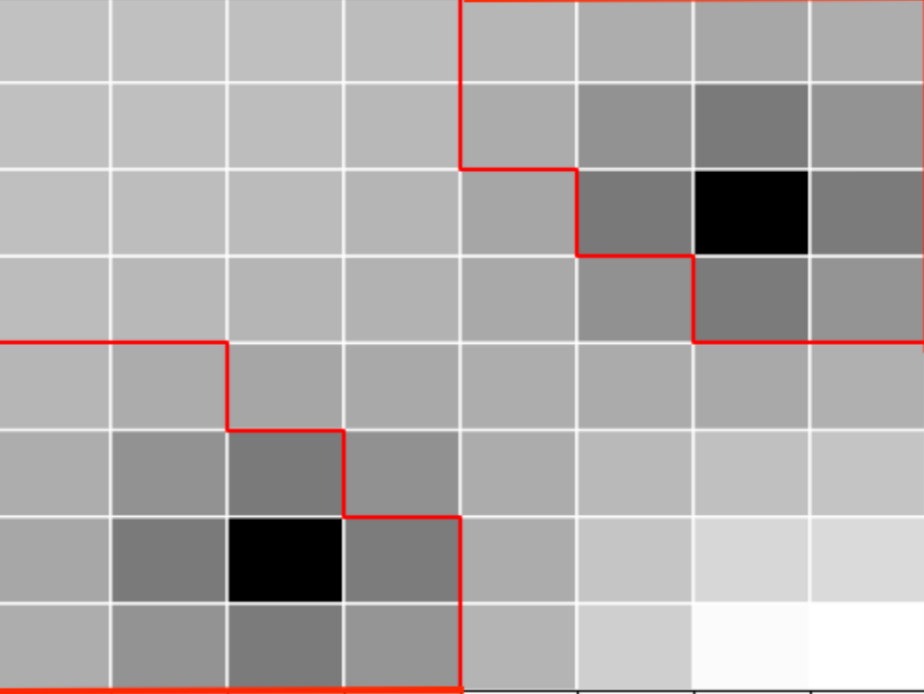}
	\caption{Reassigned reward mapping after applying safety property specification $\Phi$ with upper bound probability 0.25 in $8\times8$ Grid World Environment}
	\label{fig:initial grid environment}
\end{figure}

Also, there is a PCTL property specification $\Phi$ used for checking the safety requirement and preventing the agents from entering the unsafe area. Because if we are looking at the reward mapping in the initial Grid World environment, there is only little difference between the safe and unsafe cell states, this fact would mislead agents which are following the decision rule derived from the expert demonstrated trajectories to enter unsafe area. So it's important to reassign the reward mapping and then capture both of the goal cell states and unsafe area. This issue can be solved by applying PCTL property specification $\Phi$, which  will reassign the reward mapping of the , capture both of the goal states and unsafe area, such as what's shown in figure 5.2. 

Therefore, we defined a upper bound probability $p*$ of reaching the unsafe state within joint steps($t$) = $64\times64$(4096). Recalling from the section \ref{chapter2.6}, the property specification $\Phi$ is defined as:
\begin{equation} \label{property sepcification}
\Phi ::= P_{\leq p*}\{true \cup^{\leq t} unsafe\}
\end{equation}

By giving the prerequisite setting of our example, now we come to our actual experiment. In our experiment process, we considered 2 cases: (i) Not applying the safety property specification checking to the expert demonstrated trajectories, (ii) Applying the the safety property specification checking to the expert demonstrated trajectories.
\begin{enumerate}
\item If we do not apply the safety property specification to the expert demonstrated trajectories, then the probability of the 2 agents to enter unsafe area would be (i) 0.92 and (ii) 0.87 within 4096 joint steps $t$ by following the learnt decision rule $\hat{\pi}$ . 
\item If we are following the expert demonstrated trajectories with checking the safety property specification in figure 5.1 and set the upper bound probability $p*$ to 0.25, then based on our experiment, the 2 agents' probability of entering the unsafe area is 0.34 and 0.26 within 4096 joint steps $t$ in the entire learning process by following the learnt decision rule $\hat{\pi}$ . 
\end{enumerate}

So, based on our example experiment result, it's clear that by applying our algorithm, it actually can  lower probability of the agents entering the unsafe area in the Multi-Agent Apprenticeship Learning.


\subsection{Scalability Evaluation}
\label{chapter5.4}
Finally, we evaluated the scalability of work in the Grid Wrold environment. Table 1 shows the average runtime per iteration for the main components of our work as the size of the Grid World increases. 
\begin{enumerate}
\item The first column indicates the size of the Grid World.
\item The second columns indicates  the joint cell state space.
\item The third column indicates the average runtime that decision rule iteration would take for computing an optimal decision rule $\hat{\pi*}$ for a known joint reward function
\item The forth column indicates the average runtime that decision rule iteration would take for computing the expected features $\mu$ for a known decision rule $\hat{\pi}$. 
\item The fifth column indicates the average runtime of verifying the PCTL formula using PRISM \cite{Moldovan2012}.
\item The sixth column indicates the average runtime that generating a counterexample using COMICS \cite{Jansen2012}.
\end{enumerate}

\begin{center}
\begin{tabular}{ |p{2.5cm}|p{2.5cm}|p{2.5cm}|p{2.5cm}|p{2.5cm}|p{2.5cm}| }
\hline
\multicolumn{6}{|c|}{\textbf{Table 1 - Average runtime per iteration in seconds}} \\
\hline
\textit{Grid Size}  & \textit{Joint Cell States Num.} & \textit{Compute Decision rule} $\hat{\pi}$ & \textit{Compute State Feature} $\mu$& \textit{Model Checking} & \textit{Compute Counterexamples} \\
\hline
$3\times3$ & 81 &0.08 & 0.08 & 1.32 & 0.073\\
$8\times8$ & 4096 & 6.01 & 24.13 & 21.19 & 2.86\\
$16\times16$ & 65536 & 703.05 & 2709.01 & 2413.82 & 538.11\\
\hline
\end{tabular}
\end{center}

\newpage
\section{Conclusion}
\label{chapter6}

In this project, we have successfully made extension to the Single-Agent Apprenticeship Learning to Multi-agent Apprenticeship Learning by extending the game environment from Markov Decision Process to Markov Game and designed the extended novel framework. This section's main contents are written in following subsections:
\begin{enumerate}
\item[(\ref{chapter6.1})] Summary of Achievement
\item[(\ref{chapter6.2})] Future Work
\end{enumerate}

\subsection{Summary of Results}

\label{chapter6.1}
In this project, we first reviewed the prerequisite knowledge for building the ground knowledge basis in section \ref{chapter2} by introducing the knowledge of Reinforcement Learning Basics, Markov Game, Inverse Reinforcement Learning, and PCTL model checking. After the introduction in section \ref{chapter2}, where we should have enough background knowledge to understand the following contents of our project, we reviewed the framework of Single-Agent Safety-Aware Apprenticeship Learning in section \ref{chapter3}, which is the key part that we are making extension on.

After section \ref{chapter3}, we first introduced the knowledge of Markov Game and Multi-Agent Reinforcement Learning, and developed our extended theoretical algorithm in the following subsections in the section \ref{chapter4} and talked about the key differences between our framework and the previous work.

In the section \ref{chapter5}, we finalized our project experiment evaluation and tackled the problem of finding a decision rule $\hat{\pi}$ , in which all individual agent's policy would satisfy the PCTL property specification $\Phi$ and have low probability of entering the unsafe area in the Multi-Agent Learning environment.   

Throughout this project, we have successfully met our project goal:
\begin{enumerate}
\item extracted the reward functions in the Multi-agent Inverse Reinforcement Learning system based on the joint expert's demonstrated trajectories,
\item extended the learning learning framework from Single-Agent case to Multi-Agent case, and,
\item evaluated our framework performance in section \ref{chapter5}. 
\end{enumerate}

\subsection{Future Work}
\label{chapter6.2}
\begin{enumerate}
\item Due to the consideration of our work's complexity, we are currently setting that the agents in the Markov Game environment are able to take actions independently without having interactions. In the future, we are aiming to add constrains to the actions of the agents so that they would have interactions in the Markov Game environment. 
\item Currently, we are not considering any kind of game scenario, such as adversarial or cooperative game scenario. In the future, we are heading to consider both game scenarios and make them as extensions to our project. 
\item Due to the time related issue, we currently can only manually make the extension to the original work\cite{ZhouLi2018} from Single-Agent MDP to Two-Agent Markov Game. In the future, we are aiming to create a tool which is able to automatically create N-Agent Markov Game based on the original work.
\end{enumerate}

\newpage
\section{Acknowledgements}

\begin{enumerate}
\item First, I want to give a lot of thanks to Dr. Francesco Belardinelli and Borja Gonzalez for the theoretical and technical supports throughout this project.  If I didn't receive the supports from you, it's impossible for me to learn so much from finishing this project,.
\item Second, I want to thank to my family and friends for always mentally supporting me at the time when I wanted to give up. 
\item Finally, I want to thank to Imperial College London for providing me such precious research opportunity and project experience.
\end{enumerate}

\newpage
\bibliographystyle{abbrv}
\bibliography{sample.bib}

\end{document}